\documentclass{article}

\usepackage{microtype}
\usepackage{graphicx}
\usepackage{subcaption}
\usepackage{booktabs} 

\usepackage{hyperref}

\usepackage[preprint]{icml2026}

\usepackage{enumitem}
\usepackage{amsmath}
\usepackage{amssymb}
\usepackage{mathtools}
\usepackage{amsthm}
\usepackage[capitalize,noabbrev]{cleveref}

\usepackage{bm}
\usepackage{ifthen}
\usepackage{tabularx}
\usepackage{resizegather}
\usepackage{array}
\usepackage{multirow}

\theoremstyle{plain}
\newtheorem{theorem}{Theorem}

\theoremstyle{definition}
\newtheorem{definition}{Definition}

\theoremstyle{remark}

\icmltitlerunning{On Language Drift during RLVR Post-Training} 

\def\basemodel{{\pi}}
\def\datadist{{D}}
\def\sftdist{{\datadist^{(\textit{SFT})}}}
\def\rldist{{\datadist^{(\textit{RL})}}}
\def\nldist{{P^{(\textit{HL})}}}
\def\exprwd{{\mathcal{R}}}

\newcommand{\condist}[2][]{(#1\hspace{1mm}|\hspace{1mm}#2)}
\newcommand{\kldiv}[2][]{D_\textit{KL}(#1\hspace{1mm}||\hspace{1mm}#2)}
\newcommand{\sftmodel}[1][]{\pi^{(\textit{SFT})}_{#1}}
\newcommand{\rlmodel}[1][]{\pi^{(\textit{RL})}_{#1}}

\newcommand{\todop}[1]{{\color[HTML]{F5252C} \textbf{(TODO: #1)}}}
\newcommand{\todocite}[1]{\todop{CITE\ifthenelse{\equal{#1}{}}{}{ #1}}}

\begin{document}

\twocolumn[
  \icmltitle{On Language Drift during RLVR Post-Training}



  \icmlsetsymbol{equal}{*}

  \begin{icmlauthorlist}
    \icmlauthor{Michael Sullivan}{yyy}
    \icmlauthor{Alexander Koller}{yyy}
  \end{icmlauthorlist}

  \icmlaffiliation{yyy}{Department of Language Science and Technology, Saarland University, Saarbr{\"u}cken, Germany}

  \icmlcorrespondingauthor{Michael Sullivan}{msullivan@lst.uni-saarland.de}

  \icmlkeywords{Machine Learning, ICML}

  \vskip 0.3in
]



\printAffiliationsAndNotice{}  

\begin{abstract}

Recent advances in LLM reasoning models\textemdash driven primarily by the paradigm of post-training via reinforcement learning with verifiable reward (RLVR)\textemdash have enabled them to accomplish impressively complex tasks. However, in parallel with their rising capabilities, LLMs have increasingly displayed signs of language drift in their chains of thought (CoTs): unusual, non-standard, and seemingly nonsensical language use. Although it is well-documented\textemdash and can potentially impair CoT monitorability\textemdash the causes of language drift are thus far poorly understood. In this paper, we identify the conditions under which language drift occurs: we prove theoretically that RLVR optimization pressure permits unbounded language drift, while supervised fine-tuning does not. We then show empirically that language drift specifically arises during RLVR on novel reasoning tasks\textemdash i.e.\ when the target behavior cannot be drawn out of the base model. Finally, we prove that it is not possible to constrain language drift without constraining expected reward, suggesting that CoT monitorability cannot be improved without harming performance during RLVR post-training at the frontier.  

\end{abstract}

\section{Introduction}
\label{sec_intro}

Recent, drastic gains in the reasoning capabilities of frontier models have enabled LLM-based agents to solve increasingly complex, long-horizon coding tasks \citep{huang2026deepswe}, solve long-standing mathematical problems \citep[e.g.][]{alon2026remarks}, etc. These impressive and rapid improvements have been driven primarily by scaling reinforcement learning \citep[RL;][etc.]{openai2024o1,guo2025deepseek}\textemdash in particular, reinforcement learning with verifiable reward (RLVR), which optimizes for deterministic, outcome-based rewards such as final-answer correctness.

\begin{figure}[t]
\begin{center}
\includegraphics[width=82mm]{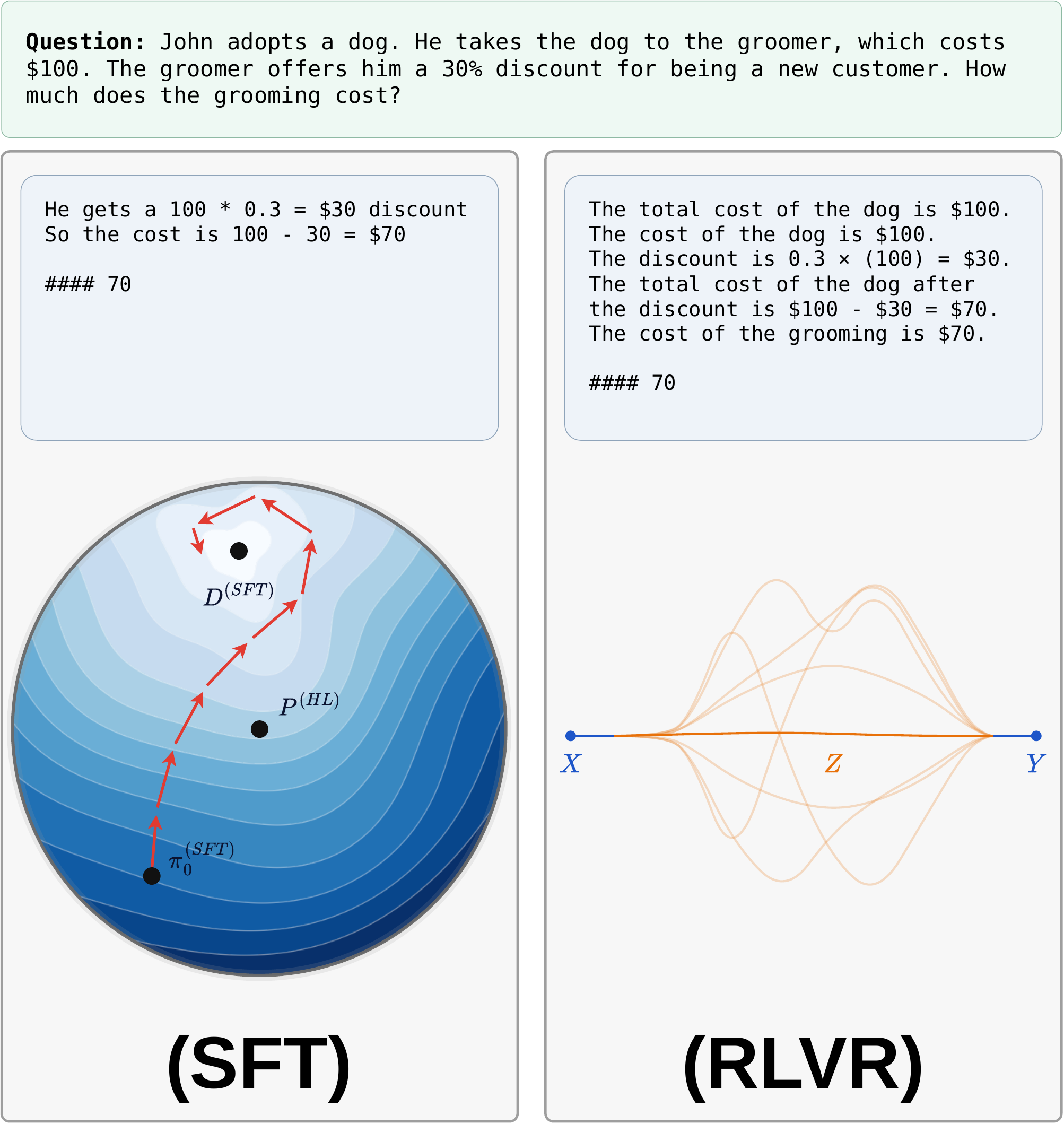}
\end{center}
\caption{
SFT (left) on prompt/CoT/answer triples $(x,z,y)$ does not permit a model $\sftmodel[]$ to drift arbitrarily far from a human language distribution $\nldist$ while minimizing cross-entropy with a target distribution $\sftdist$. RLVR (right) places no optimization pressure on the CoT $z$, allowing arbitrary language drift away from natural language in the reasoning trace. The example completions in this figure were drawn from Llama-3.2-1B models trained via SFT and RLVR (respectively) on GSM8K. 
}
\label{fig_fig_1}
\end{figure}

However, frontier LLMs have simultaneously been displaying increasing degrees of language drift: novel linguistic behavior that is non-standard and/or ungrammatical\textemdash and occasionally illegible\textemdash to human readers \citep{lazaridou-etal-2020-multi}. As an example, the RLVR-trained model in Figure \ref{fig_fig_1} has undergone language drift, affecting not only the style of its reasoning trace, but also the interpretation of the reference (meaning) of the terms that it uses. In particular, it conflates the cost of the dog\textemdash which is not relevant or stated\textemdash with the cost of the grooming.

\begin{figure}[t]
\begin{center}
\includegraphics[width=82mm]{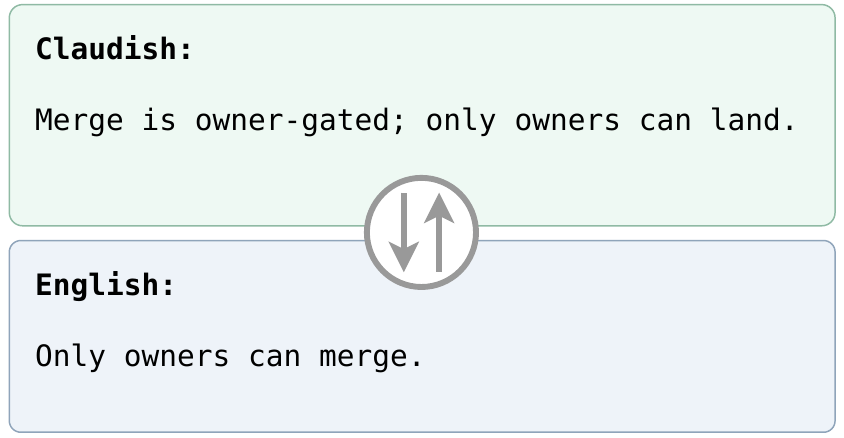}
\end{center}
\caption{``Claudish''\textemdash language in the idiosyncratic style of Anthropic's Claude models\textemdash translated to its plain-English equivalent. This example was taken from the \href{https://programasweights.com/claudish}{ProgramAsWeights Claudish-to-English translator}.} 
\label{fig_en_to_cl}
\end{figure}

Language drift is naturally frustrating to users, who struggle to parse the atypical constructions and idiosyncratic vocabulary use displayed by LLM agents. The ubiquity and severity of this problem are such that attempted workarounds have become frequent topics of online discussion: skills\footnote{e.g.\ \href{https://github.com/danyuchn/asd-ste100-skill}{https://github.com/danyuchn/asd-ste100-skill}.} instructing models to use only ASD-STE100 Simplified Technical English, ``Claudish-to-English translators'' (see Figure \ref{fig_en_to_cl}), etc. 

\begin{figure}[t]
\begin{center}
\includegraphics[width=82mm]{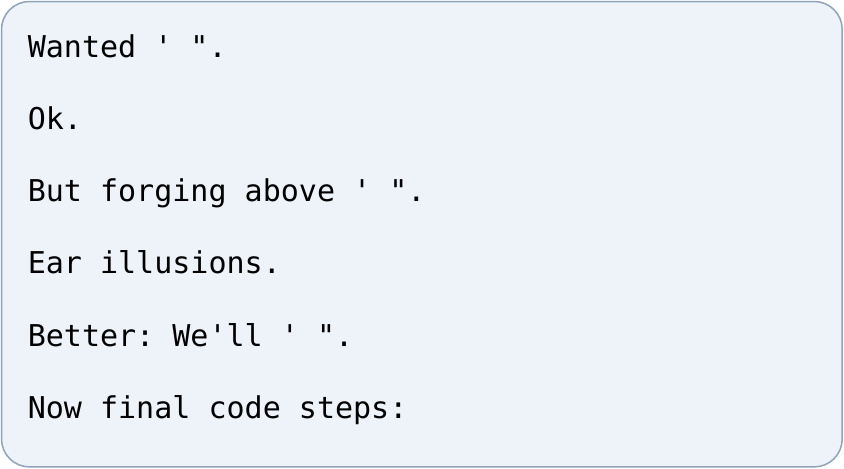}
\end{center}
\caption{Example of illegible internal reasoning uncovered during the \href{https://metr.org/evaluations/gpt-5-report/}{METR evaluation of OpenAI's GPT-5}.}
\label{fig_metr}
\end{figure}

Beyond user frustration with difficult prose, more extreme degrees of language drift (see e.g.\ Figure \ref{fig_metr}) pose a more serious threat in the form of decreased chain of thought (CoT) monitorability \citep{baker2025monitoring,korbak2025chain}: illegible CoTs\textemdash either to humans or other LLMs\textemdash render models' internal reasoning processes more opaque, leading to difficulty in determining intent and predicting future actions. This consequence of language drift is particularly worrisome in light of safety concerns due to recent security incidents involving unintended LLM agent behavior\footnote{e.g. the \href{https://openai.com/index/hugging-face-incident-and-the-road-ahead/}{OpenAI Hugging Face hacking incident}.}.

Despite its prevalence and potential consequences, the reasons for language drift remain poorly understood. Although RLVR has been suggested as a potential culprit 
\citep{korbak2025chain}, it has thus far been unclear whether language drift can occur purely due to RLVR optimization pressure. In this paper, we answer this question in the affirmative: namely, we prove theoretically that \textbf{RLVR training alone permits unbounded language drift, while supervised fine-tuning (SFT) prevents language drift} beyond a fixed, finite upper bound (see Figure \ref{fig_fig_1}; Section \ref{sec_theory}).

We additionally prove in Section \ref{sec_theory} that \textbf{constraining language drift necessarily constrains reward}, in the sense that limiting the degree of language drift that an LLM can undergo during RLVR training limits the maximum expected reward that the model can attain. This result supports arguments that maintaining CoT monitorability may not be possible without hindering downstream performance \citep[e.g.][]{baker2025monitoring}.


We then investigate the precise cause of language drift. We argue that language drift is simply a special case of novel behavior\textemdash namely, novel \textit{linguistic} behavior. Given that novel behavior arises during RLVR when \textit{behavior sharpening}\textemdash drawing target reasoning behavior out of the base model\textemdash is not possible \citep{liu2026prorl}, we hypothesize that \textbf{language drift occurs as a side-effect of novel behavior discovery during RLVR on novel tasks}. In Section \ref{sec_experiments}, we provide empirical evidence in support of this hypothesis, demonstrating that when behavior sharpening is not possible, models display greater degrees of language drift during RLVR than during SFT.

We further provide evidence that different RLVR training runs starting from the same model\textemdash and trained on the same dataset\textemdash develop distinct linguistic behavior: i.e.\ \textbf{each RLVR training run undergoes \textit{unique} language drift}. A potential consequence of this finding is that even models trained on the same tasks may not be able to monitor the CoTs of an LLM that has undergone sufficient language drift. 

Taken together, these findings paint a troubling picture of the future balance between LLM safety and capability. Namely, language drift occurs during RLVR on novel reasoning tasks\textemdash precisely those tasks which must be conquered to advance the frontier\textemdash and cannot be constrained without harming expected performance. This conclusion raises the question: as frontier models advance to human-level capability and beyond, will it remain possible to monitor their reasoning traces in order to verify alignment?

\section{Related Work}
\label{sec_related_work}

\paragraph{Language Drift during RL.}
LLMs often exhibit nonsensical linguistic patterns under RLHF-style \citep{ouyang2022training} optimization pressure: for example, models can learn to repeatedly generate the same token when using rewards derived from a neural sentiment-analysis classifier \citep{korbak2022reinforcement}. While this certainly qualifies as language drift in the narrowest sense, we exclude such reward-hacking-like phenomena from our analysis, and focus instead on the development of novel\textemdash but internally coherent\textemdash linguistic behavior. 
 
To that end, \citet{mordatch2018emergence} study the emergence of communicative behavior resembling language during multi-agent RL with a limited action space and simple, RNN-based agents initialized from scratch. They find that goal-based reward provides sufficient optimization pressure for these agents to invent language-like, compositional communication codes over an abstract set of symbols. Similarly, \citet{lazaridou-etal-2020-multi} observe language drift\textemdash including shifting word meanings\textemdash when training LSTM-based \citep{hochreiter1997long} language models to play referential games via multi-agent RL.

These studies demonstrate that RL optimization pressure can cause linguistic invention in settings targeting communication ability. Our results in Section \ref{sec_experiments} extend these findings to more general and widely used reasoning settings, and to modern transformer architectures and RLVR algorithms.

\paragraph{Illegible Reasoning Traces.}
\citet{guo2025deepseek} report that DeepSeek-R1-Zero exhibits ``poor readability and language mixing''\footnote{It is unclear to what degree language mixing represents genuine language drift in the sense of Definition \ref{def_ld}, as language mixing is also well-documented in bilingual human language users \citep[in the form of code-switching; see e.g.][]{poplack1980sometimes}.}, in particular switching between English and Chinese. Although adding a language-specific penalty improves readability, it harms performance: this finding is consistent with our theoretical result in Section \ref{sec_theory_sub_ld_is_necessary} (Theorem \ref{thm3}) that language drift cannot be constrained without negatively impacting expected reward.

\citet{jose2025reasoning} finds that frontier reasoning models sometimes output illegible and (seemingly) nonsensical CoTs within otherwise valid reasoning traces. Furthermore, removing these illegible portions of the reasoning traces considerably harms performance, suggesting that although the text may be nonsensical to human readers, it is still useful for\textemdash and therefore legible to\textemdash the LLM. \citet{panfilov2026stealing} similarly find human-illegible text in frontier models' internal reasoning traces.

Our theoretical and empirical findings in Sections \ref{sec_theory}-\ref{sec_experiments} shed light on these results demonstrating language drift in SoTA, RLVR-trained reasoning models. In particular, our results suggest that this observed language drift arose during RLVR post-training on novel reasoning tasks. 

\paragraph{Inducing Novel Linguistic Behavior.}
\citet{vaduguru-etal-2026-success} use DPO-style preference optimization \citep{rafailov2023direct} to train LLMs to perform ad-hoc linguistic convention formation in multi-turn dialog settings. Although these authors deliberately pressure LLMs to develop unique linguistic behavior, their objective is efficiency in agent-human communication, and their training method therefore cannot induce (nor does it target) arbitrary language drift. 

\citet{zhu2026large} use prompting-based methods to compress text into human-illegible form\textemdash without substantially impacting its LLM-legibility\textemdash in order to condense models' context windows without information loss. While this method merely elicits language drift potential already present in the model (as it relies purely on prompting), it demonstrates that LLMs do not necessarily need to operate over human-legible language.

\section{Preliminaries}
\label{sec_prelim}

We focus our analysis in this work on text-based reasoning LLMs: models that produce a CoT and an answer, given a prompt. We deliberately exclude multimodal models for the sake of analytical tractability, and leave an investigation of such models to future work. 

As such, we treat an LLM as a conditional probability distribution $\basemodel$ over $\mathcal{Z}\times\mathcal{Y}$ given $\mathcal{X}$\textemdash i.e.\ over pairs $(z,y)$ of CoTs $z$ and answers $y$, given a prompt $x$.

\subsection{Quantifying Language Drift}
\label{sec_prelim_sub_quant_ld}

In this work, we quantify the language drift $\operatorname{LD}(\basemodel)$ of a model $\pi$ relative to a ground-truth human language distribution $\nldist$. We define $\nldist$ to be a joint distribution over $\mathcal{X}\times\mathcal{Z}\times\mathcal{Y}$, in order to permit direct comparison with $\basemodel$.

Computing $\nldist$ is of course in practice intractable, and we do not aim to do so here: our analysis simply treats $\nldist$ as an unknown\textemdash but fixed\textemdash point in probability space. We permit the possibility that $\nldist$ is derived from an underlying, context-conditioned distribution $\mathbb{P}^{(\textit{HL})}$: for example, in the context of math reasoning tasks, we may assume that $\nldist=\mathbb{P}^{(\textit{HL})}\condist[\text{\textendash}]{\textit{task}=\textit{math-reasoning}}$.

Concretely, we model $\operatorname{LD}(\basemodel)$ as KL divergence with $\nldist$. Because the LLM itself has no control over the prompts provided to it, we specifically calculate $\operatorname{LD}(\basemodel)$ using expected KL divergence over distributions conditioned on prompts $x$ drawn from the marginal $P^{(\textit{HL})}_\mathcal{X}$. \\

\begin{definition}[Language Drift]
\label{def_ld}
    Given a reference language distribution $\nldist$ and an LLM $\basemodel$, the \textbf{language drift $\bm{LD(\basemodel)}$} is the expected KL divergence:
    \begin{align*}
        \operatorname{LD}(\basemodel)=\underset{x\sim P^{(\textit{HL})}_\mathcal{X}}{\mathbb{E}}[D_\textit{KL}(\nldist\condist[\text{\textendash}]{x}\hspace{1mm}||\hspace{1mm}\basemodel\condist[\text{\textendash}]{x})]
    \end{align*}
\end{definition}

\citet{shenfeld2026razor} measure catastrophic forgetting in a checkpoint $\basemodel$ as the expected KL divergence $\mathbb{E}_{x\sim\tau}[\kldiv[\basemodel_0]{\basemodel}]$ with the base model $\basemodel_0$ over prompts $x$ drawn from a task distribution $\tau$: analogously, we may view $\operatorname{LD}(\basemodel)$ as the degree to which $\basemodel$ has catastrophically forgotten how to produce human-like language in its CoT. 

\subsection{Modeling LLM Training}
\label{sec_prelim_sub_training}

In order to theoretically analyze the respective behavior of SFT and RLVR training with respect to language drift, it is first necessary to formally model the notions of SFT and RLVR training runs.

We model an SFT run as an \textit{SFT Checkpoint Family} (Definition \ref{def_sft}): a sequence of probability distributions (checkpoints) with monotonically non-increasing cross-entropy relative to a target distribution $\sftdist$. Here, $\sftdist$ is interpreted as the underlying distribution from which the instances in a given SFT dataset are drawn. \\

\begin{definition}[SFT Checkpoint Family]
\label{def_sft}
    Given a distribution $\sftdist$ over $\mathcal{X}\times\mathcal{Z}\times\mathcal{Y}$, an \textbf{SFT checkpoint family over $\bm{\sftdist}$} is a sequence $\sftmodel[0],\dots,\sftmodel[n]$ of conditional distributions over $\mathcal{Z}\times\mathcal{Y}$ given $\mathcal{X}$ such that for all $1\leq k\leq n$:
    \begin{equation*}
        \operatorname{CE}(\sftdist,\sftmodel[k])\leq\operatorname{CE}(\sftdist,\sftmodel[k-1])
    \end{equation*}
\end{definition}

where $\operatorname{CE}(\sftdist,\basemodel)$ denotes the expected cross-entropy over prompts drawn from the marginal $D^{(\textit{SFT})}_\mathcal{X}$:
\begin{equation}
\label{eq_ce}
\begin{split}
    \operatorname{CE}(\sftdist,\pi)=\underset{x\sim D^{(\textit{SFT})}_\mathcal{X}}{\mathbb{E}}\left[-\sum_{(z,y)}\sftdist\condist[z,y]{x}\cdot\operatorname{ln}\basemodel\condist[z,y]{x}\right]
\end{split}
\end{equation}

For the sake of simplicity, we restrict our analysis to RLVR training in the case of outcome-only reward based on final-answer correctness. Concretely, for a given $(x,y)$ prompt/answer pair, we assume a reward function that assigns a reward of 1 to the submitted final answer $\hat{y}$ if $\hat{y}=y$, and assigns a reward of 0 otherwise. 

In this setting, we may express the expected reward of a model $\basemodel$ on a distribution $\rldist$ over prompt/answer pairs $(x,y)$ as in Equation \ref{eq_exprwd}, where $\basemodel\condist[y]{x}=\sum_z\basemodel\condist[y]{x,z}\cdot\basemodel\condist[z]{x}$:
\begin{equation}
\label{eq_exprwd}
    \exprwd(\rldist,\basemodel)=\mathbb{E}_{(x,y)\sim\rldist}[\basemodel\condist[y]{x}]
\end{equation}


Analogously to the SFT case (Definition \ref{def_sft}), we model an RLVR training run as an \textit{RLVR Checkpoint Family} (Definition \ref{def_rl}): a sequence of checkpoints with monotonically non-decreasing expected reward relative to a target distribution $\rldist$. Similarly, we interpret $\rldist$ as the underlying distribution from which a given RLVR dataset is drawn.

\begin{definition}[RLVR Checkpoint Family]
\label{def_rl}
    Given a distribution $\rldist$ over $\mathcal{X}\times\mathcal{Y}$, an \textbf{RLVR checkpoint family over $\bm{\rldist}$} is a sequence $\rlmodel[0],\dots,\rlmodel[n]$ of conditional distributions over $\mathcal{Z}\times\mathcal{Y}$ given $\mathcal{X}$ such that for all $1\leq k\leq n$:

    \begin{equation*}
        \exprwd(\rldist,\rlmodel[k])\geq\exprwd(\rldist,\rlmodel[k-1])
    \end{equation*}
\end{definition}

\subsection{Assumptions}
\label{sec_prelim_sub_assumptions}

We make a few mild assumptions throughout this work in order to simplify the proofs of Theorems \ref{thm_rl}-\ref{thm3}. First, we assume that $\nldist$, $\sftdist$, and all models $\pi$ have full support on $\mathcal{X}\times\mathcal{Z}\times\mathcal{Y}$, and that $\rldist$ has full support on $\mathcal{X}\times\mathcal{Y}$. We assume further that $\rldist\condist[\text{\textendash}]{x}$ is not uniform for every $x\in\mathcal{X}$.

We additionally assume a finite vocabulary size and finite prompt, CoT, and answer lengths $\ell_\mathcal{X},\ell_\mathcal{Z},\ell_\mathcal{Y}$ (respectively). This assumption reflects the reality of LLMs with finite token vocabularies and finite context windows. All proofs are valid for any choice of $\ell_\mathcal{X},\ell_\mathcal{Z},\ell_\mathcal{Y}\geq1$.

\section{When Does Language Drift Occur?}
\label{sec_theory}

\subsection{RLVR Permits Language Drift, SFT Does Not}
\label{sec_theory_sub_theorems}

We prove that SFT optimization pressure results in a finite upper bound on the degree of possible language drift that a model can undergo (Theorem \ref{thm_sft}), while RLVR optimization does not carry such an upper bound (Theorem \ref{thm_rl}). Although \citet{korbak2025chain} informally hypothesize that RLVR may no longer incentivize legibility, Theorems \ref{thm_rl}-\ref{thm_sft} represent, to the best of our knowledge, the first formal proofs that RLVR permits unbounded language drift, while SFT does not.

Informally, RLVR permits unbounded language drift because this training regimen maximizes only the probability of producing a correct answer $y$ for a given prompt $x$. Critically, the conditional $\rlmodel[n]\condist[\mathcal{Z}]{\mathcal{X}}$ can drift arbitrarily far from $\nldist\condist[\mathcal{Z}]{\mathcal{X}}$, without negatively altering the expected reward: it is precisely this potentially unbounded KL drift in the variable $\mathcal{Z}$ which drives potentially unbounded language drift across the board. 

\begin{theorem}
\label{thm_rl}

For any choice of $\nldist$, $\rlmodel[0]$, and $\rldist$ under the assumptions in Section \ref{sec_prelim_sub_assumptions}:
\begin{equation*}
    \underset{\rlmodel[0],\dots,\rlmodel[n]}{\sup}\operatorname{LD}(\rlmodel[n])=\infty
\end{equation*}

Where $\underset{\rlmodel[0],\dots,\rlmodel[n]}{\sup}$ denotes the supremum over all RLVR checkpoint families over $\rldist$ initialized with $\rlmodel[0]$.

\end{theorem}
\begin{proof}
    Appendix \ref{app_proofs_sub_thm2}.
\end{proof}

\begin{figure}[t]
\begin{center}
\includegraphics[width=82mm]{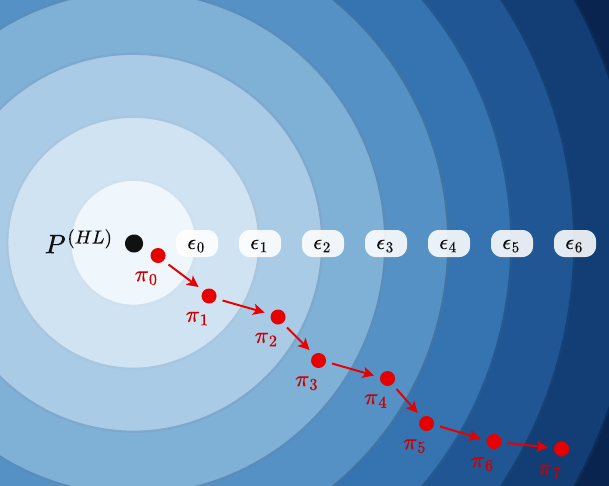}
\end{center}
\caption{For every radius $\epsilon_n$ around $\nldist$, let $\basemodel_n$ be the highest-scoring model on $\rldist$ such that $\operatorname{LD}(\basemodel_n)\leq\epsilon_n$. By Theorem \ref{thm3}, there is always some $\basemodel_{n+1}$ with $\operatorname{LD}(\basemodel_{n+1})>\epsilon_n$ that scores higher than $\basemodel_n$ on $\rldist$ (illustrated here by an arrow $\basemodel_n\rightarrow\basemodel_{n+1}$).}
\label{fig_kl_radii}
\end{figure}

On the other hand, SFT optimization constrains language drift, because an SFT dataset is drawn from some fixed distribution $\sftdist$: although $\sftdist$ may vary substantially from the human language distribution $\nldist$ (e.g.\ API/tool-use training data), $\kldiv[\nldist]{\sftdist}$ is \textit{constant}.

As such, there is necessarily a fixed, finite upper bound on the language drift for any SFT checkpoint $\sftmodel[n]$ derived from a base model $\sftmodel[0]$ (modeled as an SFT checkpoint family $\sftmodel[0],\dots,\sftmodel[n]$; see Definition \ref{def_sft}).

\begin{theorem}
\label{thm_sft}

For any choice of $\nldist$, $\sftmodel[0]$, and $\sftdist$ under the assumptions in Section \ref{sec_prelim_sub_assumptions}, there exists $\gamma<\infty$ such that:
\begin{equation*}
    \underset{\sftmodel[0],\dots,\sftmodel[n]}{\sup}\operatorname{LD}(\sftmodel[n])\leq\gamma
\end{equation*}

Where $\underset{\sftmodel[0],\dots,\sftmodel[n]}{\sup}$ denotes the supremum over all SFT checkpoint families over $\sftdist$ initialized with $\sftmodel[0]$.

\end{theorem}
\begin{proof}
    Appendix \ref{app_proofs_sub_thm1}.
\end{proof}

As a consequence of Theorem \ref{thm_sft}, no SFT run can exceed a fixed degree $\gamma$ of drift from the human language distribution $\nldist$ while still minimizing cross-entropy with the target distribution $\sftdist$.

\begin{figure}[t]
\begin{center}
\includegraphics[width=82mm]{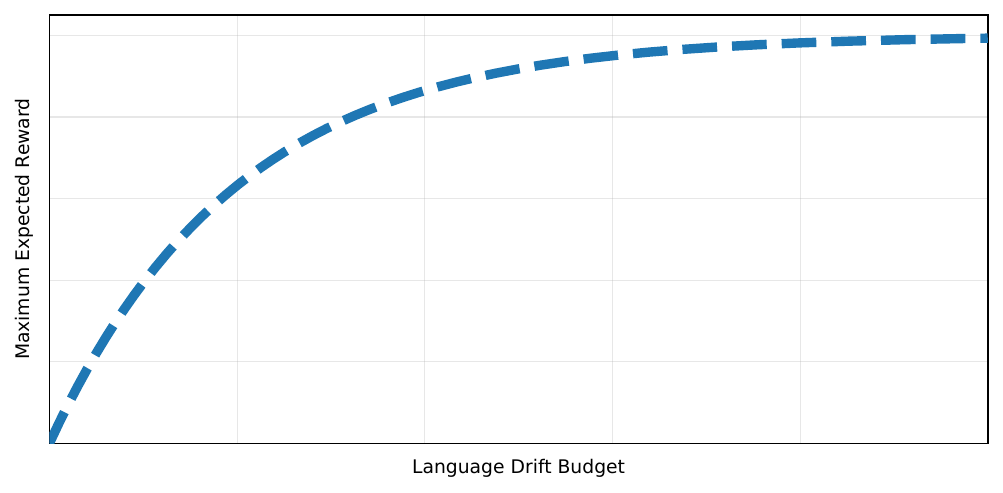}
\end{center}
\caption{As a consequence of Theorem \ref{thm3}, the maximum possible expected reward that a model can achieve on $\rldist$ asymptotically approaches $\sup_\basemodel\exprwd(\rldist,\basemodel)$ as language drift approaches $\infty$.}
\label{fig_expected_reward_vs_drift}
\end{figure}

\subsection{Constraining Language Drift Constrains Expected Reward}
\label{sec_theory_sub_ld_is_necessary}

We now prove that language drift cannot be constrained during RLVR training without limiting expected reward. Specifically, Theorem \ref{thm3} states that for every language drift radius $\epsilon$ around $\nldist$, there is a larger radius $\delta>\epsilon$ which permits a greater expected reward than $\epsilon$ (see Figure \ref{fig_kl_radii}). 

Conceptually, this is because the optimal policy\textemdash the distribution $\basemodel^*$ that maximizes the expected reward defined in Equation \ref{eq_exprwd}\textemdash concentrates all of its conditional probability mass on a single point: the most-likely answer $y^*=\operatorname{argmax}_y\rldist\condist[y]{x}$ for a given prompt $x$. As such, for every non-optimal model $\basemodel$, there exists a model $\basemodel'$ such that $\basemodel\condist[y^*]{x}<\basemodel'\condist[y^*]{x}$ and $\operatorname{LD}(\basemodel)<\operatorname{LD}(\basemodel')$, derived by interpolating between $\basemodel$ and $\basemodel^*$.

\begin{theorem}
\label{thm3}

Let $\nldist$, $\rldist$ be distributions conforming to the assumptions in Section \ref{sec_prelim_sub_assumptions}. Then for all $\epsilon\geq0$, there exists $\delta>\epsilon$ such that:
\begin{equation*}
    \max_{\basemodel\in\mathcal{B}(\delta)}\exprwd(\rldist,\basemodel)>\max_{\basemodel\in\mathcal{B}(\epsilon)}\exprwd(\rldist,\basemodel)
\end{equation*}

Where $\mathcal{B}(\epsilon)=\{\basemodel\hspace{1mm}|\hspace{1mm}\operatorname{LD}(\basemodel)\leq\epsilon\}$.
 
\end{theorem}
\begin{proof}
    Appendix \ref{app_proofs_sub_thm3}.
\end{proof}

Although the KL divergence driving language drift in Theorem \ref{thm3} can result entirely from shifting probability mass in the answer distribution $\basemodel\condist[\mathcal{Y}]{\mathcal{X},\mathcal{Z}}$\textemdash and therefore does not necessitate change in the CoT distribution $\basemodel\condist[\mathcal{Z}]{\mathcal{X}}$\textemdash altering $\basemodel\condist[\mathcal{Y}]{\mathcal{X},\mathcal{Z}}$ while holding $\basemodel\condist[\mathcal{Z}]{\mathcal{X}}$ constant still undeniably constitutes language drift. 


To illustrate this point, consider a frontier-level task $\rldist$ that humans struggle to solve: in such a case, $\nldist\condist[\mathcal{Z},\mathcal{Y}]{\mathcal{X}}$ by definition concentrates probability mass on incorrect trace/answer pairs $(z_-,y_-)$. If $\basemodel$ were to increase expected reward without substantial drift in $\basemodel\condist[\mathcal{Z}]{\mathcal{X}}$, then $\basemodel$ would necessarily produce pairs of incorrect traces $z_-$ followed by correct answers $y_+$, which implies that the interpretation of the trace $z_-$ has drifted substantially from its original meaning. 

It follows from Theorem \ref{thm3} that during RLVR post-training, as maximum expected reward approaches $\sup_\basemodel\exprwd(\rldist,\basemodel)$, minimum possible language drift approaches infinity (see Figure \ref{fig_expected_reward_vs_drift}). As a consequence, it is not possible to prevent language drift while also maximizing expected reward: constraining language drift to a fixed degree necessarily constrains the maximum possible expected reward that can be achieved. This result may therefore explain existing empirical evidence indicating that linguistic regularization harms performance on reasoning tasks \citep[see e.g.][]{guo2025deepseek}.

\subsection{Hypothesis: Novel Tasks Induce Language Drift}
\label{sec_theory_sub_hypothesis}



Theorem \ref{thm_rl} implies that RLVR permits a given training run to undergo an arbitrary degree of language drift, while continuing to maximize expected reward. Theorem \ref{thm3} goes further, and states that language drift \textit{must} occur to maximize expected reward.

These theorems do not, however, establish the degree of language drift that must occur during \textit{practical} RLVR training: for some task/model pairs, it is likely the case that only a minimal amount of language drift is required to achieve a practically acceptable expected reward. In other words, we do not establish the (likely task-dependent) steepness of the curve in Figure \ref{fig_expected_reward_vs_drift}.

As such, it is entirely possible that some SFT run undergoes a \textit{higher} degree of language drift than an RLVR training run on the same task. The point is rather that the \textit{maximum possible} language drift that an RLVR training run can undergo is unbounded, and therefore greater than the maximum of an SFT run. 


This is to say that while RLVR provides a setting in which language drift \textit{can} arise, it does not in itself \textit{cause} language drift to occur. In fact, RL's Razor \citep{shenfeld2026razor} posits that RL prefers to find solutions with minimal KL divergence to the base model, which seemingly suggests that RLVR tends to not lead to language drift. This aligns with recent findings that RLVR simply sharpens distributions, i.e.\ only draws out reasoning patterns already present in the base model \citep{chen2026does}.

However, \citet{liu2026prorl} find that LLMs can in fact develop novel behavior under RLVR, when training on novel reasoning tasks. We argue that\textemdash absent optimization pressure on the CoT $z$\textemdash those same pressures driving novel behavior development similarly increase the likelihood that novel linguistic behavior (i.e.\ language drift) arises: we hypothesize that \textbf{language drift occurs as a side-effect of novel behavior discovery during RLVR over novel reasoning tasks}. 

Conversely, we argue that language drift does not occur when the targeted reasoning task is not novel to the model: by definition, a non-novel task can be solved by drawing behavior out of the base model (behavior sharpening). In this case, RL's Razor implies that the lowest-KL path\textemdash behavior sharpening\textemdash will be preferred, and therefore no meaningful language drift will occur.

\section{Experiments}
\label{sec_experiments}

We first aim to demonstrate empirically our theoretical findings in Section \ref{sec_theory_sub_theorems} that RLVR permits greater language drift than SFT. Our second objective is to validate our hypothesis laid out in Section \ref{sec_theory_sub_hypothesis} that novel tasks in particular induce language drift in the context of RLVR training.

\begin{figure*}[t]
\begin{center}
\includegraphics[width=0.8\textwidth]{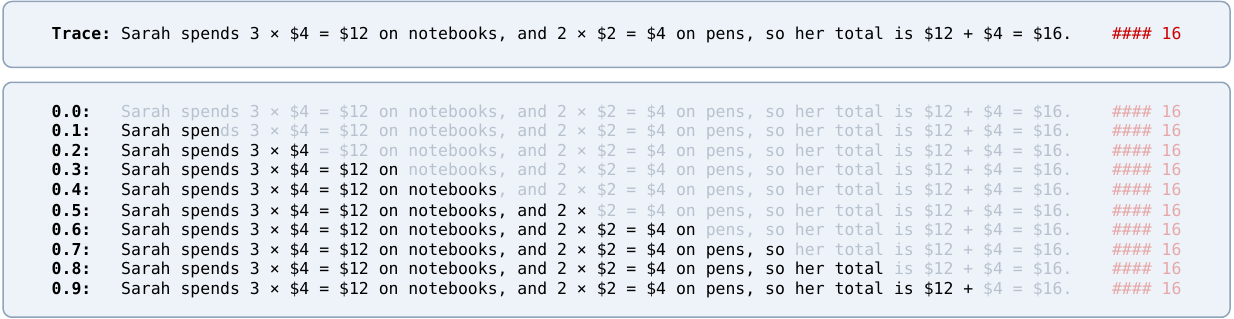}
\end{center}
\caption{Illustration of our trace legibility experiment. The target model $\pi$ first generates a full reasoning trace and answer $(z,y)\sim\basemodel\condist[\text{\textendash}]{x}$ (top). We then produce ten evenly spaced partial traces from $z$ (bottom), and for each partial trace, prompt the instruct models to guess the submitted answer $y$ based on the partial trace (with 0.0 as a control). In this highly legible example, the final answer $y=16$ can be guessed with high probability from partial traces $\geq$0.6.}
\label{fig_trace_handoff}
\end{figure*}

To that end, we study three models using the GSM8K math reasoning dataset \citep{cobbe2021training}: gemma-3-1b-pt\footnote{\href{https://huggingface.co/google/gemma-3-1b-pt}{https://huggingface.co/google/gemma-3-1b-pt}} \citep{team2025gemma}, Llama-3.2-1B\footnote{\href{https://huggingface.co/meta-llama/Llama-3.2-1B}{https://huggingface.co/meta-llama/Llama-3.2-1B}} \citep{grattafiori2024llama}, and Qwen2.5-1.5B\footnote{\href{https://huggingface.co/Qwen/Qwen2.5-1.5B}{https://huggingface.co/Qwen/Qwen2.5-1.5B}} \citep{yang2025qwen}. We strictly limit our experiments to base models, in order to avoid potential confounding factors arising from dataset contamination during mid- and/or post-training. In addition, the use of poorly performing base models allows us to use a simple\textemdash and therefore computationally inexpensive\textemdash task to emulate a difficult, novel task, while furthermore having annotated reasoning traces available for SFT tuning.

We make all code used in these experiments available on GitHub\footnote{\href{https://github.com/coli-saar/language-drift}{https://github.com/coli-saar/language-drift}}.

\subsection{Methodology}
\label{sec_experiments_sub_method}

To evaluate the difference in language drift between SFT and RLVR, we trained the base models using both training methods. We employed the DAPO \citep{yu2026dapo} variant of GRPO \citep{shao2024deepseekmath} as our RLVR implementation.

\subsubsection{Training and Validation}
\label{sec_experiments_sub_method_sub_training}

For all three models, we first performed RLVR and SFT hyperparameter grid searches to identify the best-performing configuration for each training method (further details are located in Appendix \ref{app_method_sub_training}). We then replicated the optimal-hyperparameter run for each model and training method across five random seeds, for a total of 30 training runs. 

During training and validation, the models were prompted with two fixed few-shot examples (see Figure \ref{fig_fs_prompt} in the Appendix). During validation and RLVR training, we limited the models to 256 generated tokens. Answers were submitted using the format ``\#\#\#\#\{ANSWER\}'', and we extracted the final answer detected in the model's response to compute reward/correctness. 

\paragraph{Iso-Performance Checkpoint Selection.}
For each training run, we selected for evaluation the highest-scoring checkpoint on a validation set of 128 withheld examples. However, the Llama and Gemma SFT runs consistently outperformed their RLVR counterparts. To control for any possible confounding relationship between final validation performance and language drift, we additionally selected the checkpoint from each SFT run with the closest validation score to that of the RLVR run for the same model and seed: we refer to these checkpoints as \textit{iso-SFT} during evaluation.

On the other hand, the Qwen RLVR checkpoints consistently outperformed SFT: we therefore also selected the checkpoint from each Qwen RLVR run with the closest validation score to that of the Qwen SFT run with the same seed (\textit{iso-RLVR}).

\paragraph{Behavior-Sharpening Detection.} 
As discussed above, we extracted the final submitted answer for validation and RLVR training reward. However, as base models, all three untrained LLMs failed to stop after emitting an answer to the prompted question; instead, they simply continued generating in-distribution question/trace/answer triples until the length limit was reached. 

As such, we computed accuracy for the base LLMs relative to both the \textit{first} and last submitted answer, in order to obtain a rough estimate of the models' propensity for behavior sharpening: if an LLM has a high first-submitted answer accuracy and a low last-submitted answer accuracy, then to achieve a high reward, it needs only learn to emit the EOS token after submitting the answer (rather than generate another question). In this case, the model clearly does not need to discover new behavior, and so RLVR can solve the task primarily through behavior sharpening. 

\subsubsection{Evaluation}
\label{sec_experiments_sub_method_sub_evaluation}

As discussed in Section \ref{sec_prelim}, we do not have access to the human language distribution $\nldist$. As such, it is not possible to directly measure language drift in these experiments. We instead use three instruct models to approximate $\nldist$: Llama-3.2-1B-Instruct\footnote{\href{https://huggingface.co/meta-llama/Llama-3.2-1B-Instruct}{https://huggingface.co/meta-llama/Llama-3.2-1B-Instruct}}, Qwen2.5-0.5B-Instruct\footnote{\href{https://huggingface.co/Qwen/Qwen2.5-0.5B-Instruct}{https://huggingface.co/Qwen/Qwen2.5-0.5B-Instruct}}, and OLMo-2-0425-1B-Instruct\footnote{\href{https://huggingface.co/allenai/OLMo-2-0425-1B-Instruct}{https://huggingface.co/allenai/OLMo-2-0425-1B-Instruct}} \citep{olmo2024}.

Although by our own arguments made throughout this paper, all three of those post-trained LLMs have themselves undergone an unknown degree of language drift from $\nldist$, it is exceedingly unlikely that three distinct models have drifted \textit{to the same point} in probability space. As such, we argue that by averaging over the three models, we can obtain a noisy triangulation of $\nldist$.

Even with this instruct-model-derived approximation of $\nldist$, computing the full KL divergence between two LLMs is still intractable. We therefore measure relative language drift using CoT legibility as a proxy: the more an external reader is able to follow the CoT of an LLM, the less language drift that LLM has undergone.  

\paragraph{Trace Legibility.}
We implement an experimental design based on those used in \citet{wang-etal-2025-chain} and \citet{roytburg2026measuring}. First, for each checkpoint $\basemodel$ selected after training in Section \ref{sec_experiments_sub_method} (i.e.\ best-performing and iso-SFT/RLVR) and each prompt $x$ in the GSM8K validation set, we sample a reasoning trace and answer $(z,y)\sim\basemodel\condist[\text{\textendash}]{x}$. We then construct ten partial traces from $z$, such that the $n^{th}$ partial trace spans the first $n/10$ of $z$ (see Figure \ref{fig_trace_handoff}). 

For each partial trace, each of the three instruct models is prompted to guess the final answer $y$, based on the reasoning produced by $\basemodel$ up to that point. By measuring the degree to which the instruct models are able to follow $z$, this experiment provides an estimate of the legibility of the CoT\textemdash and therefore an estimate of the (lack of) language drift undergone by $\basemodel$. We refer to this experiment as \textit{trace legibility}.

\begin{table}[t]
\centering
\begin{tabular}{l|cc|c|c}
\multirow{2}{*}{Model} & \multicolumn{2}{c|}{\underline{Base}} & \multirow{2}{*}{RLVR} & \multirow{2}{*}{SFT} \\
& First-Ans. & Last-Ans. & & \\
\hline
Llama & 0.06 & 0.02 & 0.23{\scriptsize$\pm$0.04} & 0.32{\scriptsize$\pm$0.02} \\
Gemma & 0.02 & 0.01 & 0.12{\scriptsize$\pm$0.01} & 0.24{\scriptsize$\pm$0.02} \\
Qwen  & 0.41 & 0.20 & 0.79{\scriptsize$\pm$0.01} & 0.63{\scriptsize$\pm$0.02} \\
\end{tabular}
\caption{GSM8K validation accuracy for all three model types. For the RLVR- and SFT-trained models, we report the mean accuracy ($\pm$ standard deviation) across all five training seeds. We additionally report first-answer accuracy for the base models, as a measure of behavior-sharpening potential. These two metrics are disjoint: cases where only one answer was submitted counted only towards last-answer score.}
\label{table_gsm8k_perf}
\end{table}

As we are purely interested in CoT legibility, we do not measure final-answer \textit{correctness} in this task, but rather final-answer \textit{agreement} with the answer $y$ submitted by $\basemodel$. In order to prevent the instruct models from simply ignoring the partial CoT and answering based solely on the question $x$, we omit $x$ itself from the instruct models' prompt: they are required to guess the final answer entirely from the fraction of the trace $z$ obtained from $\basemodel$. 

\paragraph{Self-Legibility.}
We perform an additional round of trace legibility experiments, in which we replace the instruct-model CoT readers with other models of the same type and trained with the same method (SFT/RLVR) as $\pi$. For example, if the target model $\pi$ is a Gemma model trained with RLVR, we replace the instruct models with the other four Gemma RLVR runs (i.e.\ the other four seeds). We refer to this experiment as \textit{self-legibility}.

The goal of the self-legibility experiment is to measure the \textit{regularity} (conversely, novelty) of language drift. If, for example, traces generated by an RLVR-trained model are more legible to other RLVR-trained models, we may conclude that RLVR-trained models tend to drift linguistically in the same direction in probability space. Conversely, if traces generated by an RLVR-trained model are \textit{less} legible to other RLVR-trained models, then we may conclude that the direction of RLVR-induced language drift is unpredictable.

Additional details regarding the configuration of the trace legibility and self-legibility experiments are located in Appendix \ref{app_method_sub_handoff}.

\paragraph{CoT Utility Sanity Check.}
For each training run, we confirm that the model is in fact utilizing its generated reasoning traces to produce final answers, rather than producing performative CoTs \citep{boppana2026reasoning}, invisible reasoning \citep{baherwani2026not}, etc. A discussion of our methodology and results is located in Appendix \ref{app_sanity_check}.

\subsection{Results}
\label{sec_experiments_sub_results}

GSM8K validation performance for each base, RLVR, and SFT model is given in Table \ref{table_gsm8k_perf}. Recall that for the base models, we record accuracy with respect to both the last-submitted answer\textemdash used during training for validation and RLVR reward calculation\textemdash \textit{and} the first-submitted answer, using the latter to estimate behavior-sharpening potential (see Section \ref{sec_experiments_sub_method}). 

\begin{figure}[t]
\begin{center}
\includegraphics[width=82mm]{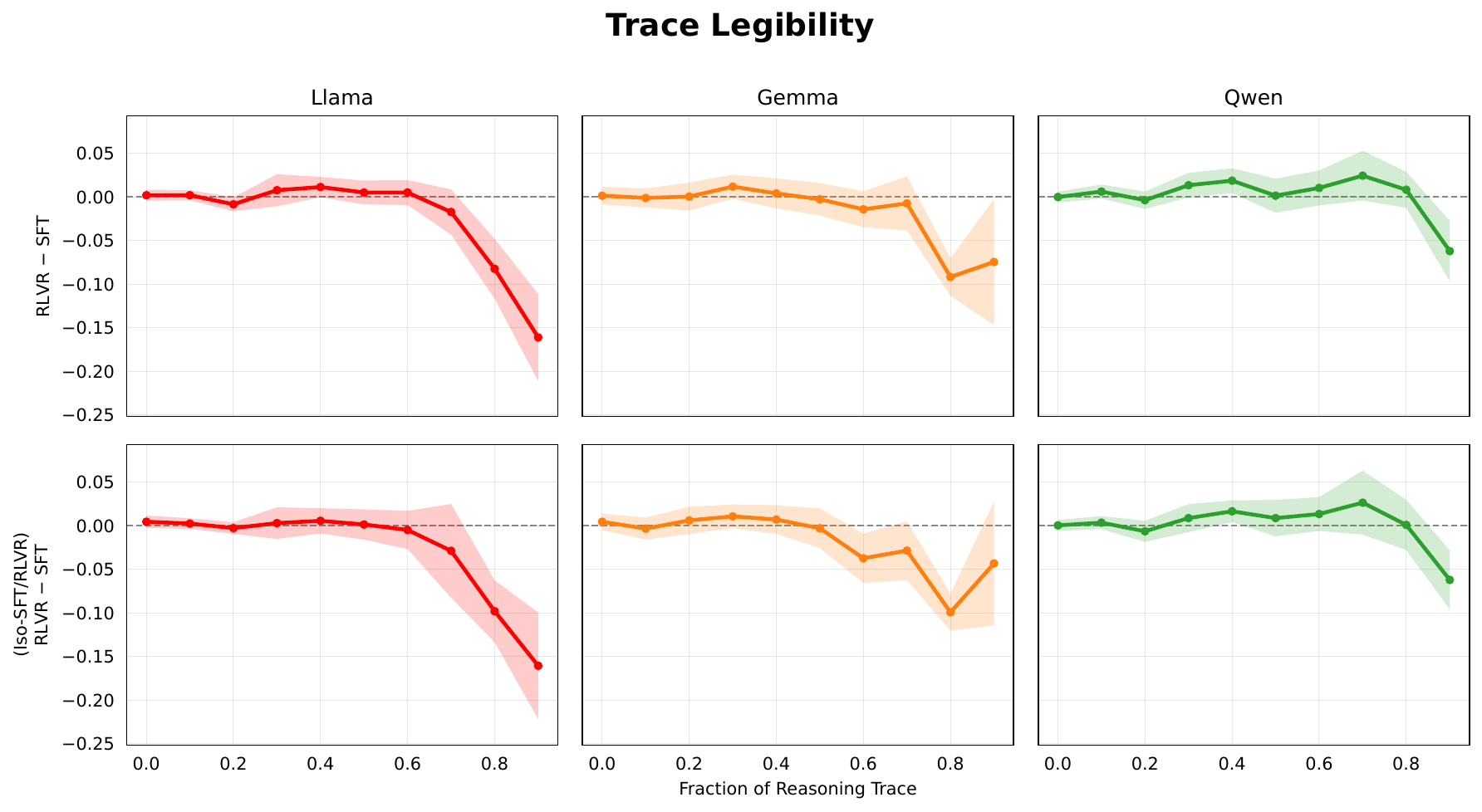}
\includegraphics[width=82mm]{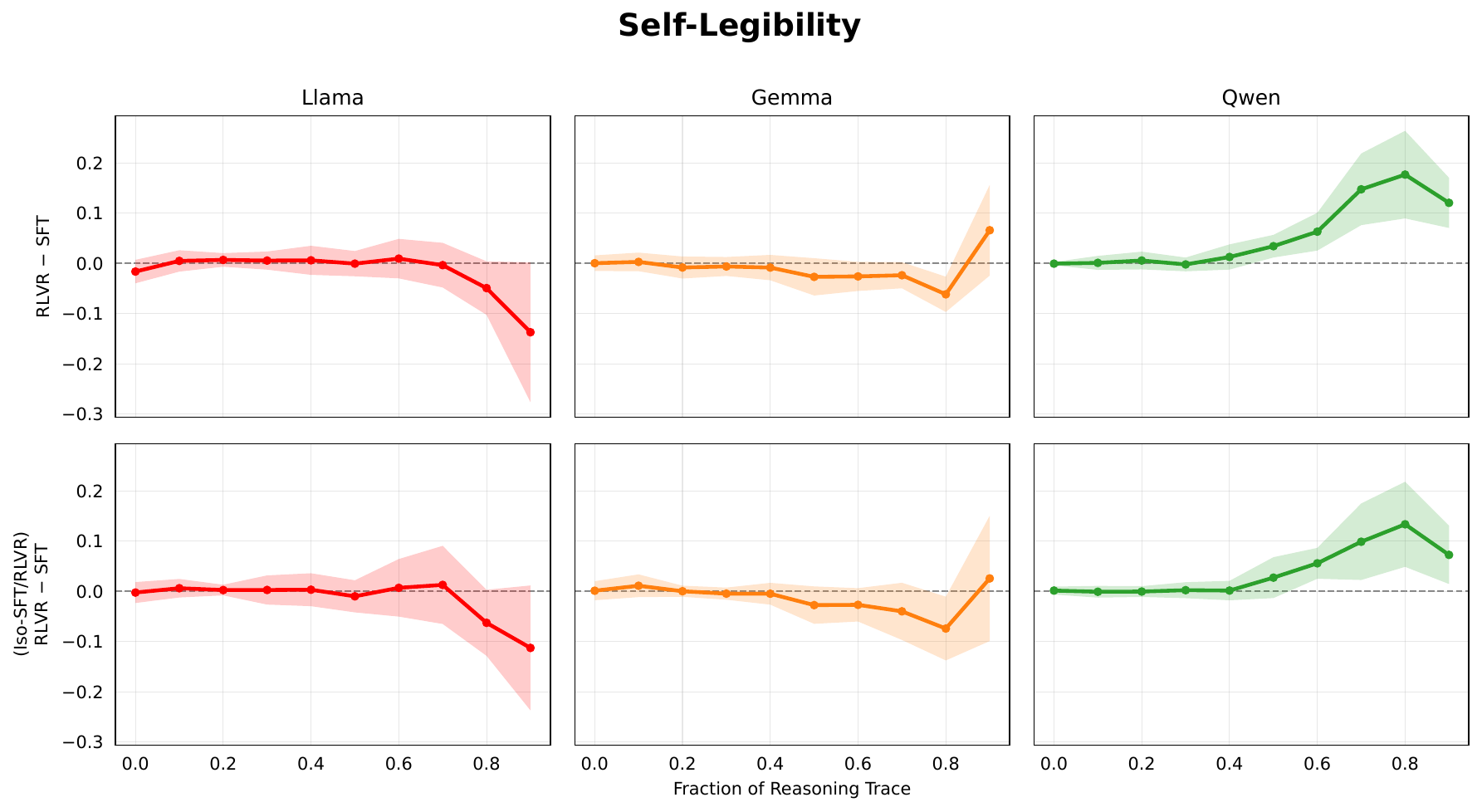}
\end{center}
\caption{Mean difference in trace legibility (first two rows) and self-legibility (last two rows) scores for the best (first and third rows) and iso-SFT/RLVR (second and fourth rows) checkpoints (95\% confidence intervals). Each point $p\in\{0.0,0.1,\dots,0.9\}$ reports the difference in the mean legibility of each trace $z$ generated by the model in question, when cut off at $\textit{len}(z)\cdot p$ tokens. Absolute values are given in Figure \ref{fig_handoff_grid_overlay} in the Appendix.}
\label{fig_handoff_grid}
\end{figure}

\paragraph{Qwen Undergoes Behavior Sharpening during RLVR.}
The base Qwen model achieves an accuracy of $\sim$0.61 when combining first-answer (0.41) and last-answer accuracy (0.20). Given that its final accuracy after RLVR is 0.79, we can conclude that around two-thirds of Qwen's RLVR accuracy gains come purely from learning to emit the EOS token after submitting an answer. 

\begin{table}[t]
\centering
\scalebox{0.93}{\begin{tabular}{l|l|ccc}
Task & Checkpoint & Llama & Gemma & Qwen \\
\hline
\multirow{2}{*}{Trace Legibility} & Best-Perf & -0.24 & -0.17 &  0.02 \\
                                & Iso-Perf  & -0.28 & -0.18 &  0.01 \\
\hline
\multirow{2}{*}{Self-Legibility}  & Best-Perf & -0.17 & -0.09 &  0.56 \\
                                & Iso-Perf  & -0.15 & -0.14 &  0.39 \\
\end{tabular}}
\caption{AUC summary of each plot in Figure \ref{fig_handoff_grid}.}
\label{table_auc}
\end{table}

This is not the case for Llama and Gemma: both models nearly triple their combined (first-answer plus last-answer) accuracy after RLVR. This is a clear indicator of novel behavior discovery. As such, we expect to see more pronounced language drift from RLVR for those models.

\paragraph{Novel Tasks Induce Language Drift.}
In Figure \ref{fig_handoff_grid}, we plot for each model the difference between mean RLVR legibility and mean SFT legibility: values less than zero indicate that the SFT traces were more legible than the RLVR traces at that fraction. We expect the differences at earlier fractions to be near zero, as a final answer is extremely difficult to guess when given less than half of the corresponding trace, regardless of whether it was generated by RLVR- or SFT-trained models. 

The plots in Figure \ref{fig_handoff_grid} are summarized by the area-under-the-curve (AUC) values in Table \ref{table_auc}. Llama and Gemma have negative values for trace legibility and self-legibility, both when comparing best-performing checkpoints and when comparing the training methods at iso-performance. This indicates that both models have higher degrees of language drift\footnote{We record anecdotal, qualitative examples of novel linguistic behavior from Llama and Gemma in Appendix \ref{app_examples}.} across the board under RLVR, which is consistent with Theorems \ref{thm_rl}-\ref{thm_sft}. That the Llama traces display higher degrees of language drift than Gemma is to be expected, as Gemma achieves much lower accuracy than Llama: when a task is too difficult, GRPO-style RL algorithms have very sparse gradients, and the LLM's weights therefore cannot drift very far from those of the base model.

As predicted by the hypothesis in Section \ref{sec_theory_sub_hypothesis}, Qwen\textemdash the only model that performs behavior sharpening during RLVR\textemdash is the only model for which RLVR exhibits virtually no language drift relative to SFT. 

\paragraph{RLVR-Induced Language Drift is Unique.}
The results of the self-legibility experiment indicate that the RLVR-trained Llama and Gemma models have a lower degree of mutual legibility than their SFT-trained counterparts. This indicates that each RLVR-trained run drifted in a different direction, which in turn indicates that each run developed unique linguistic behavior. 

Conversely, Qwen\textemdash the only model that did not undergo RLVR-induced language drift as measured by trace legibility\textemdash is the only model type for which the RLVR-trained models have a higher degree of mutual legibility than the SFT-trained models.

\section{Discussion and Conclusion}
\label{sec_conclusion}

In this paper, we presented evidence that RLVR on novel tasks induces language drift in LLMs. In Section  \ref{sec_theory}, we proved theoretically that there is a finite upper bound on the maximum possible language drift that can occur during SFT, while no such upper bound exists for RLVR. Furthermore, we proved that language drift cannot be constrained during RLVR training without constraining expected reward (Theorem \ref{thm3}).

We then demonstrated empirically that LLMs undergo a greater degree of language drift under RLVR than under SFT, when behavior sharpening is not viable\textemdash i.e.\ when the task is novel to the base model (Section \ref{sec_experiments}). We furthermore showed that such RLVR-induced drift is unique, in that models of the same type trained on different seeds suffer from decreased mutual legibility.

The implications of these findings with respect to frontier-model CoT monitorability are considerable. In particular, the conditions under which RLVR induces language drift\textemdash novel reasoning tasks\textemdash are precisely those under which models must be trained in order to advance the frontier. Furthermore, these are precisely the conditions under which RLVR is most needed, due to the difficulty of obtaining sufficient amounts of SFT data on novel reasoning environments.

It then follows from Theorem \ref{thm3} that we must accept increasing degrees of language drift in order to continue advancing the frontier with RLVR. Finally, that RLVR-induced language drift is unique\textemdash even among models initialized from the same base and trained on the same task\textemdash indicates that it is unpredictable: a given model can drift linguistically in multiple possible directions during RLVR training. This suggests that even a model trained on the same data as some target model $\basemodel$ cannot necessarily be relied upon to effectively comprehend\textemdash and therefore, monitor\textemdash the internal reasoning traces produced by $\basemodel$.

\section*{Limitations}

Due to resource constraints, we used only small, 1-1.5B models in our experiments. The use of such models naturally limits the size and difficulty of the reasoning datasets that they can be trained on: while we undeniably observe language drift in the Llama and Gemma models (see Appendix \ref{app_examples}), this drift does not reach the degree of illegibility displayed in some frontier-model internal reasoning traces (e.g.\ in Figure \ref{fig_metr}). 

Due to the restricted range of datasets that these small models can be trained on, we were only able to find one dataset (GSM8K) that both contained annotated CoTs and that could be learned through RLVR and SFT by the three base models (Llama, Gemma, and Qwen; see Section \ref{sec_experiments_sub_method}). In future versions of this manuscript, we intend to extend the results of our experiments in Section \ref{sec_experiments} with additional reasoning datasets. 

Finally, our experiments in Section \ref{sec_experiments} primarily pertain to Theorems \ref{thm_rl}-\ref{thm_sft} and our hypothesis in Section \ref{sec_theory_sub_hypothesis}: we do not include any empirical demonstration of our theoretical findings in Section \ref{sec_theory_sub_ld_is_necessary} (Theorem \ref{thm3}) that constraining language drift necessarily constrains expected reward. We likewise intend to update future versions of the manuscript with experiments to this end.

\bibliography{example_paper}
\bibliographystyle{icml2026}

\newpage
\appendix
\onecolumn

\section{Proofs}
\label{app_proofs}

\subsection{Theorem \ref{thm_rl}}
\label{app_proofs_sub_thm2}

For any $x\in\mathcal{X}$, let $z^*(x)=\operatorname{argmin}_z\nldist\condist[z]{x}$ and $y^*(x)=\operatorname{argmax}_y\rldist\condist[y]{x}$, and define $\pi^*$ to be the conditional distribution on $\mathcal{Z}\times\mathcal{Y}$ given $\mathcal{X}$ such that $\pi^*\condist[z^*(x),y^*(x)]{x}=1$.

Let $\pi_{(\text{\textendash})}$ be an infinite sequence of distributions such that for all $n\geq0$ and all $x\in\mathcal{X}$:

\begin{equation}
\label{eq_pi_seq}
    \pi_n\condist[\text{\textendash}]{x}=\frac{1}{n+1}\cdot\rlmodel[0]\condist[\text{\textendash}]{x}+\left(1-\frac{1}{n+1}\right)\cdot\pi^*\condist[\text{\textendash}]{x}
\end{equation}



As $\underset{n\rightarrow\infty}{\operatorname{lim}}\pi_n\condist[z^*(x)]{x}=1$, we have $\underset{n\rightarrow\infty}{\operatorname{lim}}\pi_n\condist[z]{x}=0$ for any $z\neq z^*(x)$. This in turn implies that $\underset{n\rightarrow\infty}{\operatorname{lim}}\pi_n\condist[z,y]{x}=0$ for all $(z,y)\in\mathcal{Z}\times\mathcal{Y}$ such that $z\neq z^*(x)$: as $\nldist\condist[z,y]{x}\neq0$ by assumption (Section \ref{sec_prelim_sub_assumptions}), it follows that $\underset{n\rightarrow\infty}{\operatorname{lim}}\operatorname{ln}(\nldist\condist[z,y]{x}/\pi_n\condist[z,y]{x})=\infty$, and therefore:

\begin{equation}
\label{eq_inf_ld}
    \underset{n\rightarrow\infty}{\operatorname{lim}}D_\textit{KL}(\nldist\condist[\text{\textendash}]{x}\hspace{1mm}||\hspace{1mm}\pi_n\condist[\text{\textendash}]{x})=\underset{n\rightarrow\infty}{\operatorname{lim}}\operatorname{LD}(\pi_n)=\infty
\end{equation}

As $\exprwd(\rldist,\pi_n)=\frac{1}{n+1}\cdot\exprwd(\rldist,\rlmodel[0])+\left(1-\frac{1}{n+1}\right)\cdot\exprwd(\rldist,\pi^*)$ by construction, it follows from the optimality of $\pi^*$ that:

\begin{equation}
\label{eq_valid_rlcf}
    \forall n\colon \exprwd(\rldist,\pi_n)\geq\exprwd(\rldist,\rlmodel[0])
\end{equation}

Now, let $\mathcal{S}=\{(\rlmodel[0],\pi_n)\hspace{1mm}|\hspace{1mm}n\geq1\}$. It follows from Equation \ref{eq_inf_ld} that:

\begin{equation}
\label{eq_sup_subset}
    \sup_{(\rlmodel[0],\pi_n)\in\mathcal{S}}\operatorname{LD}(\pi_n)=\infty
\end{equation}

By Equation \ref{eq_valid_rlcf}, each $(\rlmodel[0],\pi_n)\in\mathcal{S}$ is a valid RLVR checkpoint family initialized with $\rlmodel[0]$ (Definition \ref{def_rl}), and therefore $\mathcal{S}$ forms a subset of the set of RLVR checkpoint families initialized with $\rlmodel[0]$. It therefore follows from Equation \ref{eq_sup_subset} that:

\begin{equation*}
    \underset{\rlmodel[0],\dots,\rlmodel[n]}{\sup}\operatorname{LD}(\rlmodel[n])=\infty
\end{equation*}

Where $\underset{\rlmodel[0],\dots,\rlmodel[n]}{\sup}$ denotes the supremum over all RLVR checkpoint families over $\rldist$ initialized with $\rlmodel[0]$.

This completes the proof. 

\subsection{Theorem \ref{thm_sft}}
\label{app_proofs_sub_thm1}

Fix $x\in\mathcal{X}$. By definition, we have:

\begin{equation}
\label{eq_kl_breakdown}
\begin{split}
    D_\textit{KL}(\nldist\condist[\textit{\textendash}]{x}\hspace{1mm}||\hspace{1mm}\sftmodel[n]\condist[\textit{\textendash}]{x})&=\underset{(z,y)\sim\nldist\condist[\textit{\textendash}]{x}}{\mathbb{E}}\left[\operatorname{ln}\frac{\nldist\condist[z,y]{x}}{\sftmodel[n]\condist[z,y]{x}}\right] \\ &=\underset{(z,y)\sim\nldist\condist[\textit{\textendash}]{x}}{\mathbb{E}}\left[\operatorname{ln}\left(\frac{\nldist\condist[z,y]{x}}{\sftdist\condist[z,y]{x}}\cdot\frac{\sftdist\condist[z,y]{x}}{\sftmodel[n]\condist[z,y]{x}}\right)\right] \\
    &=\underset{(z,y)\sim\nldist\condist[\textit{\textendash}]{x}}{\mathbb{E}}\left[\operatorname{ln}\left(\frac{\nldist\condist[z,y]{x}}{\sftdist\condist[z,y]{x}}\right)+\operatorname{ln}\left(\frac{\sftdist\condist[z,y]{x}}{\sftmodel[n]\condist[z,y]{x}}\right)\right] \\ 
    &=\underset{(z,y)\sim\nldist\condist[\textit{\textendash}]{x}}{\mathbb{E}}\left[\operatorname{ln}\frac{\nldist\condist[z,y]{x}}{\sftdist\condist[z,y]{x}}\right]+\underset{(z,y)\sim\nldist\condist[\textit{\textendash}]{x}}{\mathbb{E}}\left[\operatorname{ln}\frac{\sftdist\condist[z,y]{x}}{\sftmodel[n]\condist[z,y]{x}}\right] \\
    &=D_\textit{KL}(\nldist\condist[\textit{\textendash}]{x}\hspace{1mm}||\hspace{1mm}\sftdist\condist[\textit{\textendash}]{x})+\underset{(z,y)\sim\nldist\condist[\textit{\textendash}]{x}}{\mathbb{E}}\left[\operatorname{ln}\frac{\sftdist\condist[z,y]{x}}{\sftmodel[n]\condist[z,y]{x}}\right]
\end{split}
\end{equation}

Now, define $f_x(z,y)=\operatorname{ln}\frac{\sftdist\condist[z,y]{x}}{\sftmodel[n]\condist[z,y]{x}}$, and for any real $a$, define $a^+=\max(a,0)$ and $a^-=\max(-a,0)$. Again by definition, we have:

\begin{equation}
\label{eq_abc_ineq}
\begin{split}
    &\underset{(z,y)\sim\nldist\condist[\textit{\textendash}]{x}}{\mathbb{E}}[f_x(z,y)]-\underset{(z,y)\sim\sftdist\condist[\textit{\textendash}]{x}}{\mathbb{E}}[f_x(z,y)] \\
    &=\sum_{(z,y)}\nldist\condist[z,y]{x}\cdot f_x(z,y)-\sum_{(z,y)}\sftdist\condist[z,y]{x}\cdot f_x(z,y)=\sum_{(z,y)}(\nldist\condist[z,y]{x}-\sftdist\condist[z,y]{x})\cdot f_x(z,y) \\ 
    &=\left(\sum_{(z,y)}(\nldist\condist[z,y]{x}-\sftdist\condist[z,y]{x})\cdot f_x(z,y)^+\right)-\left(\sum_{(z,y)}(\nldist\condist[z,y]{x}-\sftdist\condist[z,y]{x})\cdot f_x(z,y)^-\right) \\
    &\leq\left(\sum_{(z,y)}(\nldist\condist[z,y]{x}-\sftdist\condist[z,y]{x})^+\cdot f_x(z,y)^+\right)+\left(\sum_{(z,y)}(\sftdist\condist[z,y]{x}-\nldist\condist[z,y]{x})^+\cdot f_x(z,y)^-\right)
\end{split}
\end{equation}

Let $r_\textit{max}=\max_{(x',z,y)}\frac{\nldist(z,y\hspace{0.5mm}|\hspace{0.5mm}x')}{\sftdist(z,y\hspace{0.5mm}|\hspace{0.5mm}x')}$ and $r_\textit{min}=\min_{(x',z,y)}\frac{\nldist(z,y\hspace{0.5mm}|\hspace{0.5mm}x')}{\sftdist(z,y\hspace{0.5mm}|\hspace{0.5mm}x')}$. Then for any $(z,y)\in\mathcal{Z}\times\mathcal{Y}$, we have:

\begin{equation*}
    (\nldist\condist[z,y]{x}-\sftdist\condist[z,y]{x})^+=\left(\left(\frac{\nldist\condist[z,y]{x}}{\sftdist\condist[z,y]{x}}-1\right)\cdot\sftdist\condist[z,y]{x}\right)^+\leq(r_\textit{max}-1)\cdot\sftdist\condist[z,y]{x}
\end{equation*}

Similarly, $(\sftdist\condist[z,y]{x}-\nldist\condist[z,y]{x})^+\leq(1-r_\textit{min})\cdot\sftdist\condist[z,y]{x}$. It then follows from Equation \ref{eq_abc_ineq} that:

\begin{equation}
\label{eq_pos_neg}
\begin{split}
   \underset{(z,y)\sim\nldist\condist[\text{\textendash}]{x}}{\mathbb{E}}[f_x(z,y)&]\leq \\
    \underset{(z,y)\sim\sftdist\condist[\text{\textendash}]{x}}{\mathbb{E}}&[f_x(z,y)]+(r_\textit{max}-1)\cdot\underset{(z,y)\sim\sftdist\condist[\text{\textendash}]{x}}{\mathbb{E}}[f_x(z,y)^+]+(1-r_\textit{min})\cdot\underset{(z,y)\sim\sftdist\condist[\text{\textendash}]{x}}{\mathbb{E}}[f_x(z,y)^-]
\end{split}
\end{equation}

As $f_x(z,y)^+=f_x(z,y)+f_x(z,y)^-$, we may substitute $\underset{(z,y)\sim\sftdist\condist[\text{\textendash}]{x}}{\mathbb{E}}[f_x(z,y)^+]$ for $\underset{(z,y)\sim\sftdist\condist[\text{\textendash}]{x}}{\mathbb{E}}[f_x(z,y)]+\underset{(z,y)\sim\sftdist\condist[\text{\textendash}]{x}}{\mathbb{E}}[f_x(z,y)^-]$ in Equation \ref{eq_pos_neg} to yield:

\begin{equation*}
\begin{split}
   & \underset{(z,y)\sim\nldist\condist[\text{\textendash}]{x}}{\mathbb{E}}[f_x(z,y)]\leq \\
    &\underset{(z,y)\sim\sftdist\condist[\text{\textendash}]{x}}{\mathbb{E}}[f_x(z,y)]+(r_\textit{max}-1)\cdot\underset{(z,y)\sim\sftdist\condist[\text{\textendash}]{x}}{\mathbb{E}}[f_x(z,y)]+(r_\textit{max}-1)\cdot\underset{(z,y)\sim\sftdist\condist[\text{\textendash}]{x}}{\mathbb{E}}[f_x(z,y)^-]+(1-r_\textit{min})\cdot\underset{(z,y)\sim\sftdist\condist[\text{\textendash}]{x}}{\mathbb{E}}[f_x(z,y)^-] \\
    &=(1+(r_\textit{max}-1))\cdot\underset{(z,y)\sim\sftdist\condist[\text{\textendash}]{x}}{\mathbb{E}}[f_x(z,y)]+((r_\textit{max}-1)+(1-r_\textit{min}))\cdot\underset{(z,y)\sim\sftdist\condist[\text{\textendash}]{x}}{\mathbb{E}}[f_x(z,y)^-] \\
    &=r_\textit{max}\cdot\underset{(z,y)\sim\sftdist\condist[\text{\textendash}]{x}}{\mathbb{E}}[f_x(z,y)]+(r_\textit{max}-r_\textit{min})\cdot\underset{(z,y)\sim\sftdist\condist[\text{\textendash}]{x}}{\mathbb{E}}[f_x(z,y)^-]
\end{split}
\end{equation*}

By definition, $\underset{(z,y)\sim\sftdist\condist[\text{\textendash}]{x}}{\mathbb{E}}[f_x(z,y)]=D_\textit{KL}(\sftdist\condist[\text{\textendash}]{x}\hspace{1mm}||\hspace{1mm}\sftmodel[n]\condist[\text{\textendash}]{x})$, and so we have:

\begin{equation}
\label{eq_upto_neg}
\begin{split}
    \underset{(z,y)\sim\nldist\condist[\text{\textendash}]{x}}{\mathbb{E}}[&f_x(z,y)]\leq \\
    &r_\textit{max}\cdot D_\textit{KL}(\sftdist\condist[\text{\textendash}]{x}\hspace{1mm}||\hspace{1mm}\sftmodel[n]\condist[\text{\textendash}]{x})+(r_\textit{max}-r_\textit{min})\cdot\underset{(z,y)\sim\sftdist\condist[\text{\textendash}]{x}}{\mathbb{E}}[f_x(z,y)^-]
\end{split}
\end{equation}

It now remains to derive an upper bound on the right-hand expectation over $f_x(z,y)^-$. First, note that for all $(z,y)\in\mathcal{Z}\times\mathcal{Y}$:

\begin{equation}
\label{eq_log_switch}
    f_x(z,y)^-=\max\left(-\operatorname{ln}\frac{\sftdist\condist[z,y]{x}}{\sftmodel[n]\condist[z,y]{x}},0\right)=\max\left(\operatorname{ln}\frac{\sftmodel[n]\condist[z,y]{x}}{\sftdist\condist[z,y]{x}},0\right)
\end{equation}

Where the first equality is by definition of $f$ and $(\text{\textendash})^-$. Letting $\mathcal{S}=\{(z,y)\in\mathcal{Z}\times\mathcal{Y}\hspace{1mm}|\hspace{1mm}\sftmodel[n]\condist[z,y]{x}>\sftdist\condist[z,y]{x}\}$, it follows from Equation \ref{eq_log_switch} and the fact that $\operatorname{ln}(a)\leq a-1$ for all $a$ that:

\begin{equation*}
\begin{split}
    \underset{(z,y)\sim\sftdist\condist[\text{\textendash}]{x}}{\mathbb{E}}[f_x(z,y)^-]&=\sum_{(z,y)\in\mathcal{S}}\sftdist\condist[z,y]{x}\cdot\operatorname{ln}\left(\frac{\sftmodel[n]\condist[z,y]{x}}{\sftdist\condist[z,y]{x}}\right) \\
    &\leq\sum_{(z,y)\in\mathcal{S}}\sftmodel[n]\condist[z,y]{x}-\sftdist\condist[z,y]{x} \\ &=\sum_{(z,y)\in\mathcal{Z}\times\mathcal{Y}}\frac{1}{2}|\sftmodel[n]\condist[z,y]{x}-\sftdist\condist[z,y]{x}| \\
    &=\operatorname{TV}(\sftdist\condist[\text{\textendash}]{x},\sftmodel[n]\condist[\text{\textendash}]{x})\leq1
\end{split}    
\end{equation*}

Where $\operatorname{TV}(\sftdist\condist[\text{\textendash}]{x},\sftmodel[n]\condist[\text{\textendash}]{x})$ is the total variation distance between $\sftdist\condist[\text{\textendash}]{x}$ and $\sftmodel[n]\condist[\text{\textendash}]{x}$, which is bounded on the interval $[0,1]$. As such, we may replace the expectation over $f_x(z,y)^-$ in Equation \ref{eq_upto_neg} to yield:

\begin{equation}
\label{eq_pre_kl_shift}
    \underset{(z,y)\sim\nldist\condist[\text{\textendash}]{x}}{\mathbb{E}}[f_x(z,y)]\leq r_\textit{max}\cdot D_\textit{KL}(\sftdist\condist[\text{\textendash}]{x}\hspace{1mm}||\hspace{1mm}\sftmodel[n]\condist[\text{\textendash}]{x})+(r_\textit{max}-r_\textit{min})
\end{equation}



By assumption (Definition \ref{def_sft}), $\operatorname{CE}(\sftdist,\sftmodel[n])\leq\operatorname{CE}(\sftdist,\sftmodel[n-1])$ for all $n\geq1$: it follows that $\operatorname{CE}(\sftdist,\sftmodel[n])\leq\operatorname{CE}(\sftdist,\sftmodel[0])$ for all $n\geq0$. By definition (Equation \ref{eq_ce}), we have:

\begin{equation*}
\begin{split}
    \operatorname{CE}(\sftdist,\sftmodel[n])=\underset{x\sim D^{(\textit{SFT})}_\mathcal{X}}{\mathbb{E}}[D_\textit{KL}(\sftdist\condist[\text{\textendash}]{x}\hspace{1mm}||\hspace{1mm}\sftmodel[n]\condist[\text{\textendash}]{x})+H(\sftdist\condist[\text{\textendash}]{x})]
\end{split}
\end{equation*}

Where the entropy $H(\sftdist\condist[\text{\textendash}]{x})$ is fixed. Therefore, letting $g_n(x)=r_\textit{max}\cdot D_\textit{KL}(\sftdist\condist[\text{\textendash}]{x}\hspace{1mm}||\hspace{1mm}\sftmodel[n]\condist[\text{\textendash}]{x})+(r_\textit{max}-r_\textit{min})$, we have:

\begin{equation}
\label{eq_g_ineq}
    \forall n\geq0\colon\underset{x\sim D^{(\textit{SFT})}_\mathcal{X}}{\mathbb{E}}[g_n(x)]\leq\underset{x\sim D^{(\textit{SFT})}_\mathcal{X}}{\mathbb{E}}[g_0(x)]
\end{equation}

Now, define $r_D=\max_x \frac{P^{(\textit{HL})}_\mathcal{X}(x)}{D^{(\textit{SFT})}_\mathcal{X}(x)}$. By construction, $P^{(\textit{HL})}_\mathcal{X}(x)\leq D^{(\textit{SFT})}_\mathcal{X}(x)\cdot r_D$ for all $x\in\mathcal{X}$. Combining with Equations \ref{eq_kl_breakdown}, \ref{eq_pre_kl_shift}, and \ref{eq_g_ineq}, we have:

\begin{equation}
\label{eq_rd_g0}
\begin{split}
    \operatorname{LD}(\sftmodel[n])&=\underset{x\sim P^{(\textit{HL})}_\mathcal{X}}{\mathbb{E}}[D_\textit{KL}(\nldist\condist[\text{\textendash}]{x}\hspace{1mm}||\hspace{1mm}\sftmodel[n]\condist[\text{\textendash}]{x})] \\
    &=
    \underset{x\sim P^{(\textit{HL})}_\mathcal{X}}{\mathbb{E}}\left[D_\textit{KL}(\nldist\condist[\text{\textendash}]{x}\hspace{1mm}||\hspace{1mm}\sftdist\condist[\text{\textendash}]{x})+\underset{(z,y)\sim\nldist\condist[\text{\textendash}]{x}}{\mathbb{E}}[f_x(z,y)]\right] \\
    &\leq
    \underset{x\sim P^{(\textit{HL})}_\mathcal{X}}{\mathbb{E}}[D_\textit{KL}(\nldist\condist[\text{\textendash}]{x}\hspace{1mm}||\hspace{1mm}\sftdist\condist[\text{\textendash}]{x})+g_n(x)] \\
    &\leq\underset{x\sim D^{(\textit{SFT})}_\mathcal{X}}{\mathbb{E}}[r_D\cdot(D_\textit{KL}(\nldist\condist[\text{\textendash}]{x}\hspace{1mm}||\hspace{1mm}\sftdist\condist[\text{\textendash}]{x})+g_n(x))] \\
    &\leq\underset{x\sim D^{(\textit{SFT})}_\mathcal{X}}{\mathbb{E}}[r_D\cdot(D_\textit{KL}(\nldist\condist[\text{\textendash}]{x}\hspace{1mm}||\hspace{1mm}\sftdist\condist[\text{\textendash}]{x})+g_0(x))]
\end{split}
\end{equation}

Define $\gamma=\underset{x\sim D^{(\textit{SFT})}_\mathcal{X}}{\mathbb{E}}[r_D\cdot(D_\textit{KL}(\nldist\condist[\text{\textendash}]{x}\hspace{1mm}||\hspace{1mm}\sftdist\condist[\text{\textendash}]{x})+g_0(x))]$. That $\gamma$ is finite follows from the assumption that $\nldist$, $\sftdist$, and $\sftmodel[0]$ all have full support on $\mathcal{X}\times\mathcal{Z}\times\mathcal{Y}$ (Section \ref{sec_prelim_sub_assumptions}). As $\gamma$ does not depend on the term $n$, we have:

\begin{equation*}
    \underset{\sftmodel[0],\dots,\sftmodel[n]}{\sup}\operatorname{LD}(\sftmodel[n])\leq\gamma<\infty
\end{equation*}

This completes the proof.

\subsection{Theorem \ref{thm3}}
\label{app_proofs_sub_thm3}

Let $\pi^*$ be any conditional distribution over $\mathcal{Z}\times\mathcal{Y}$ given $\mathcal{X}$ such that:

\begin{equation}
\label{eq_pi_star}
    \forall x\colon \sum_{z\in\mathcal{Z}}\pi^*\condist[z,y^*(x)]{x}=1 
\end{equation}

Where $y^*(x)=\operatorname{argmax}_y \rldist\condist[y]{x}$. As $y^*(x)$ maximizes both $\pi^*\condist[\text{\textendash}]{x}$ and $\rldist\condist[\text{\textendash}]{x}$, $\pi^*$ is the optimal policy for $\rldist$.

Clearly, $\operatorname{LD}(\pi^*)=\infty$: pick any $x\in\mathcal{X}$ and $(z,y)\in\mathcal{Z}\times\mathcal{Y}$ such that $y\neq y^*(x)$. By assumption (Section \ref{sec_prelim_sub_assumptions}) and definition, $\nldist\condist[z,y]{x}\neq0$ and $\pi^*\condist[z,y]{x}=0$: therefore, $D_\textit{KL}(\nldist\condist[\text{\textendash}]{x}\hspace{1mm}||\hspace{1mm}\pi^*\condist[\text{\textendash}]{x})=\infty$.

Now, for any $0\leq\epsilon<\infty$, define $\pi_\epsilon$ as in Equation \ref{eq_pi_hat_epsilon}:

\begin{equation}
\label{eq_pi_hat_epsilon}   
    \pi_\epsilon=\underset{\pi\in\mathcal{B}(\epsilon)}{\operatorname{argmax}}\hspace{1mm}\exprwd(\rldist,\pi)
\end{equation}

By construction, $\operatorname{LD}(\pi_\epsilon)\leq\epsilon<\infty=\operatorname{LD}(\pi^*)$, and so it must be the case that $\pi_\epsilon\neq\pi^*$. Now, for $0<\lambda<1$, define:

\begin{equation*}
    \pi^{(\lambda)}_\epsilon\condist[\text{\textendash}]{x}=\lambda\cdot\pi^*\condist[\text{\textendash}]{x}+(1-\lambda)\cdot\pi_\epsilon\condist[\text{\textendash}]{x}
\end{equation*}

For any $0<\lambda<1$, $\exprwd(\rldist,\pi^{(\lambda)}_\epsilon)=\lambda\cdot\exprwd(\rldist,\pi^*)+(1-\lambda)\cdot\exprwd(\rldist,\pi_\epsilon)$ by construction, and therefore $\exprwd(\rldist,\pi^{(\lambda)}_\epsilon)>\exprwd(\rldist,\pi_\epsilon)$. As $\pi_\epsilon$ is the maximizer of $\mathcal{B}(\epsilon)$ by definition (Equation \ref{eq_pi_hat_epsilon}), it must be the case that $\operatorname{LD}(\pi^{(\lambda)}_\epsilon)>\epsilon$.

Fix any $\lambda\in(0,1)$, and define $\delta$ from the statement of Theorem \ref{thm3} to be $\operatorname{LD}(\pi^{(\lambda)}_\epsilon)$. As $\pi^{(\lambda)}_\epsilon$ has full support on $\mathcal{X}\times\mathcal{Z}\times\mathcal{Y}$, we have $\delta<\infty$. 

This completes the proof.


\newpage

\section{Experimental Methodology}
\label{app_method}

\subsection{Training and Validation}
\label{app_method_sub_training}

\begin{figure*}[h]
\begin{center}
\includegraphics[width=164mm]{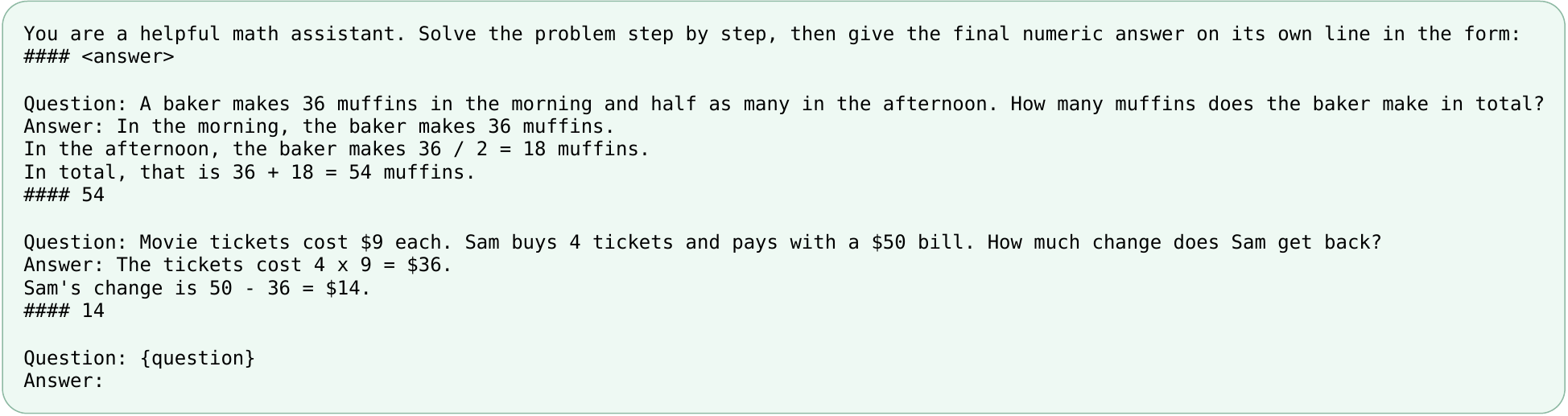}
\end{center}
\caption{Few-shot prompt used during SFT and RLVR training.}
\label{fig_fs_prompt}
\end{figure*}

\begin{table}[h]
\centering
\begin{tabular}{l|l|l}
\textbf{Training Type} & \textbf{Hyperparameter} & \textbf{Values} \\
\hline
Both & Learning rate & 1e-6, 5e-6, 1e-5, 3e-5, 5e-5 \\
\hline
\multirow{4}{*}{DAPO (RLVR)} & Group size & 64, 128 \\
 & Temperature & 0.3, 0.5 \\
 & Batch size & 128, 256 \\
 & Epochs & 1, 2 \\
\hline
\multirow{2}{*}{SFT} & Batch size & 4, 8, 16, 32 \\
 & Epochs & 1, 2, 3 \\
\end{tabular}
\caption{Hyperparameter search values by training type. We swept the same learning rates for both DAPO and SFT.}
\label{table_hp_grid_search}
\end{table}

For each model and training type, we performed a hyperparameter grid search across the values given in Table \ref{table_hp_grid_search}: DAPO required a large group size due to the low success probabilities of the base models. All models were trained with a cosine learning rate scheduler with a warmup ratio of 0.03, using the few-shot prompt in Figure \ref{fig_fs_prompt}. For the DAPO training runs, we fixed the number of update iterations ($\mu$) at 1, $\epsilon_\textit{low}$ at 0.2, and $\epsilon_\textit{high}$ at 0.28.

Each training run was performed on a single NVIDIA H100 GPU.

\subsection{Trace Legibility}
\label{app_method_sub_handoff}

\begin{figure*}[t]
\begin{center}
\includegraphics[width=82mm]{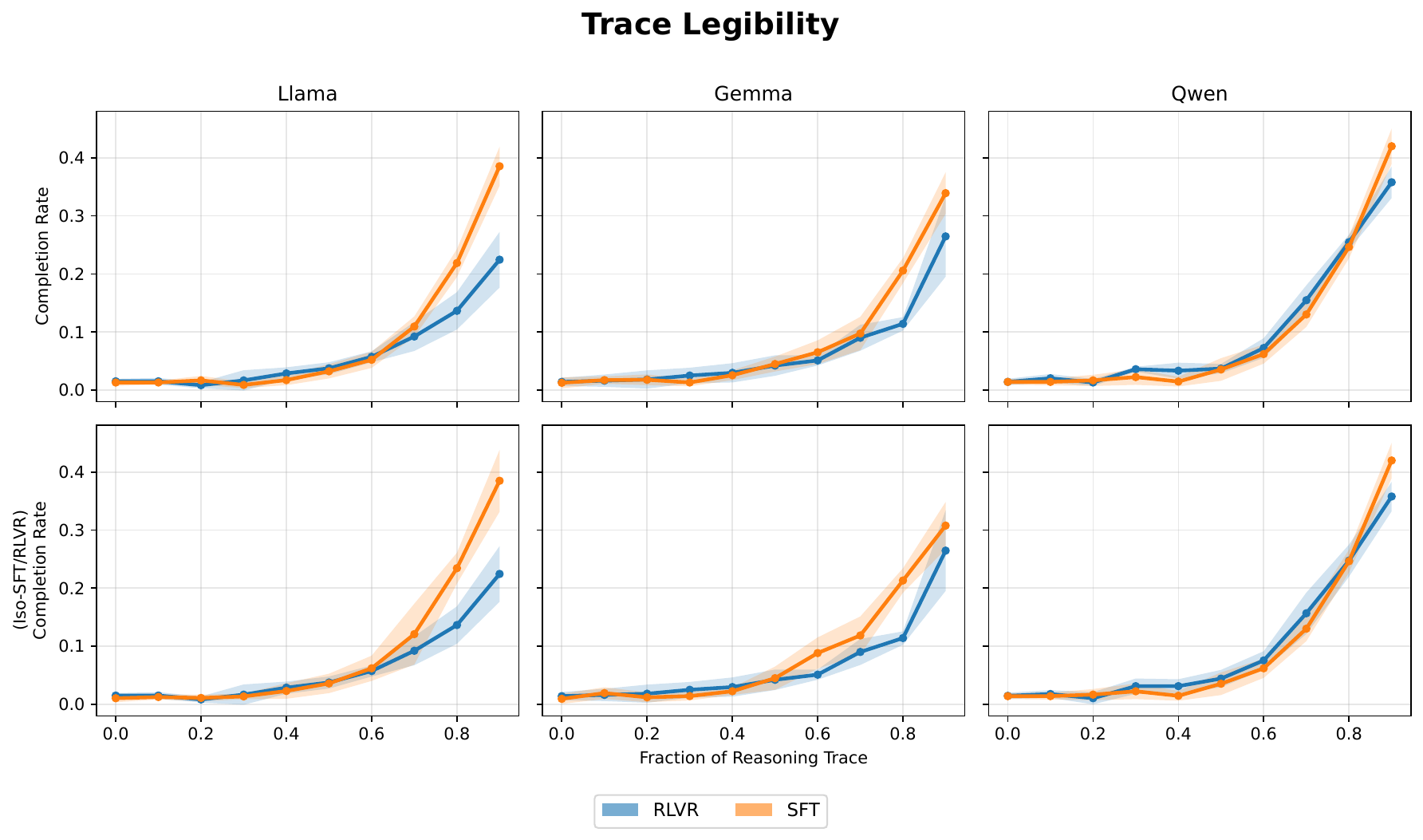}
\includegraphics[width=82mm]{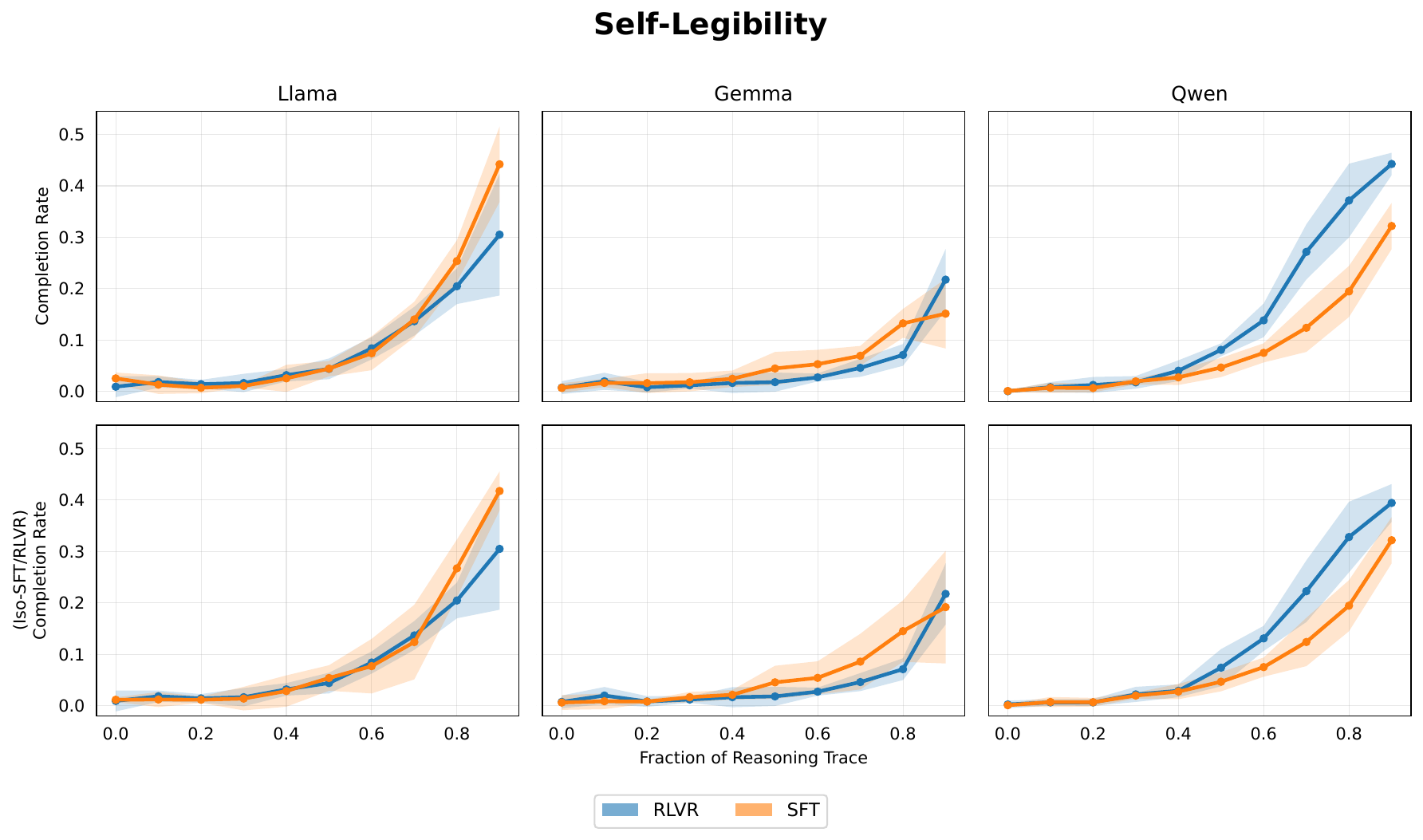}
\end{center}
\caption{Mean trace legibility (left) and self-legibility (right) scores for the best (top row) and iso-SFT/RLVR (bottom row) checkpoints (95\% confidence intervals).}
\label{fig_handoff_grid_overlay}
\end{figure*}

\begin{figure*}[t!]
\begin{center}
\includegraphics[width=164mm]{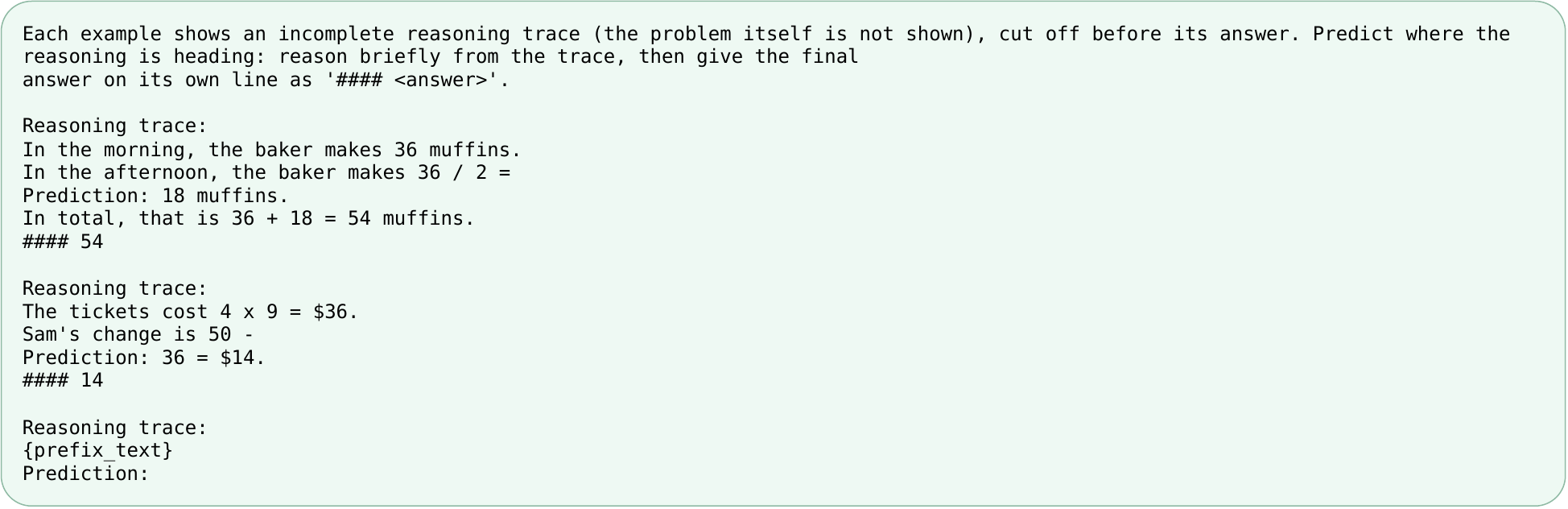}
\end{center}
\caption{Few-shot prompt used to elicit final-answer guessing in the self-legibility experiment.}
\label{fig_fs_prompt_base}
\end{figure*}

\begin{figure*}[t!]
\begin{center}
\includegraphics[width=164mm]{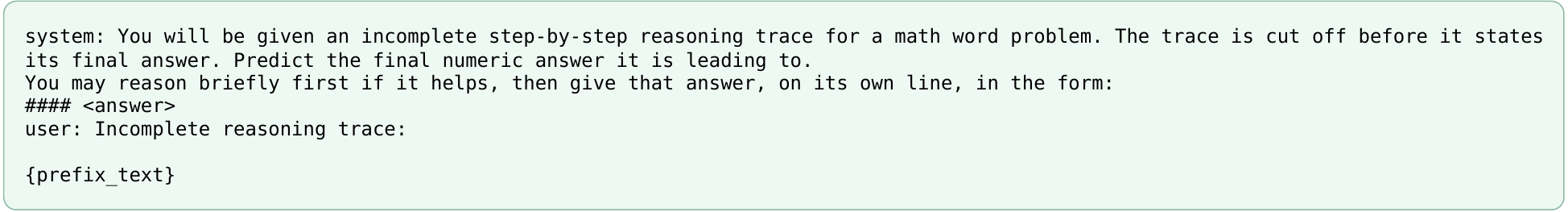}
\end{center}
\caption{Prompt used with the instruct models in the trace legibility experiment.}
\label{fig_fs_prompt_instruct}
\end{figure*}

In the trace legibility task, all instruct models were prompted using the prompt in Figure \ref{fig_fs_prompt_instruct}, and in the self-legibility task, all models were prompted using the prompt in Figure \ref{fig_fs_prompt_base}. Sampling for all models\textemdash generators and readers\textemdash was carried out deterministically.

\newpage

\hspace{0mm}

\newpage

\section{CoT Utility Sanity Check}
\label{app_sanity_check}

\begin{figure*}[h]
\begin{center}
\includegraphics[width=82mm]{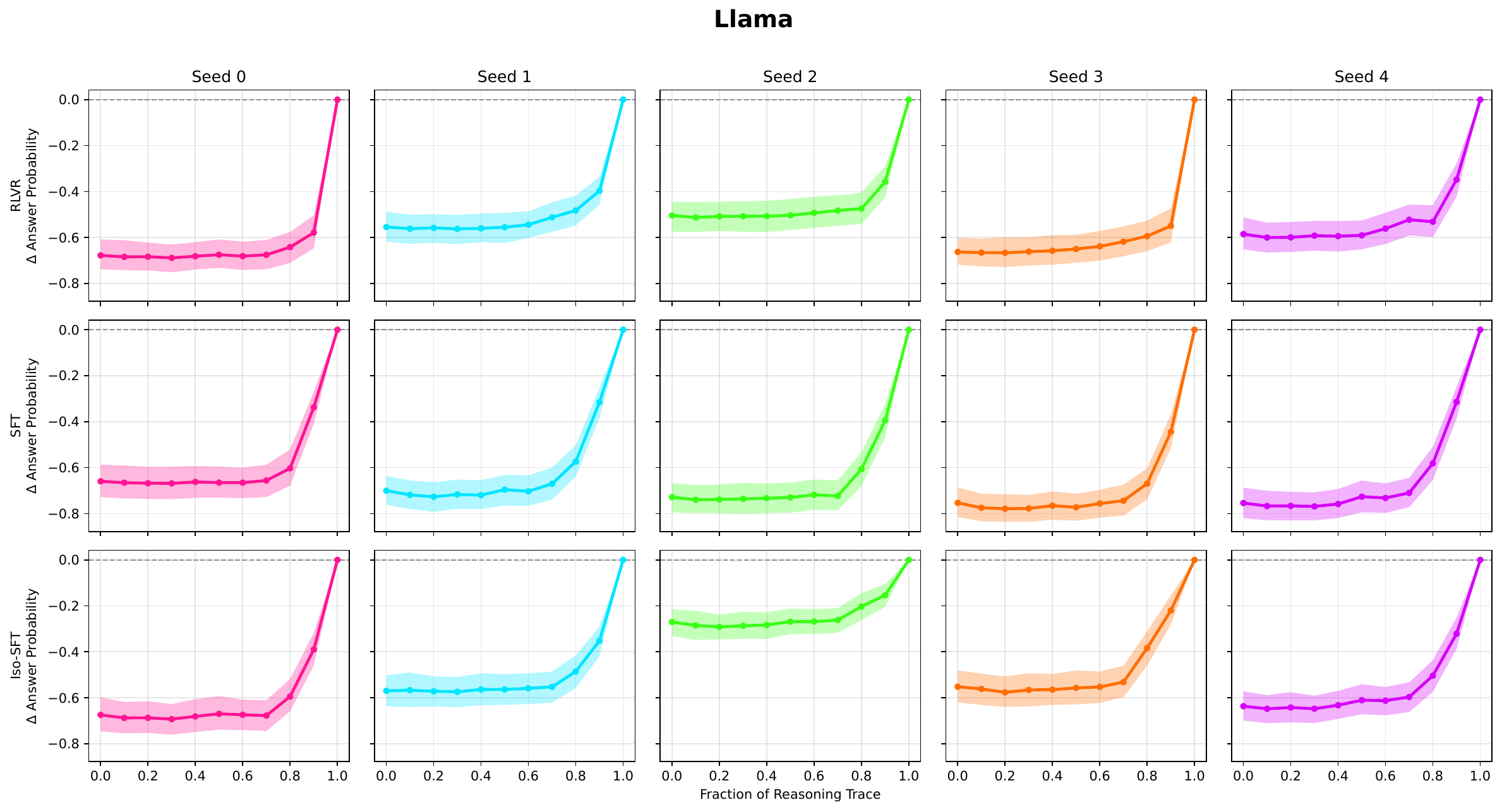}
\includegraphics[width=82mm]{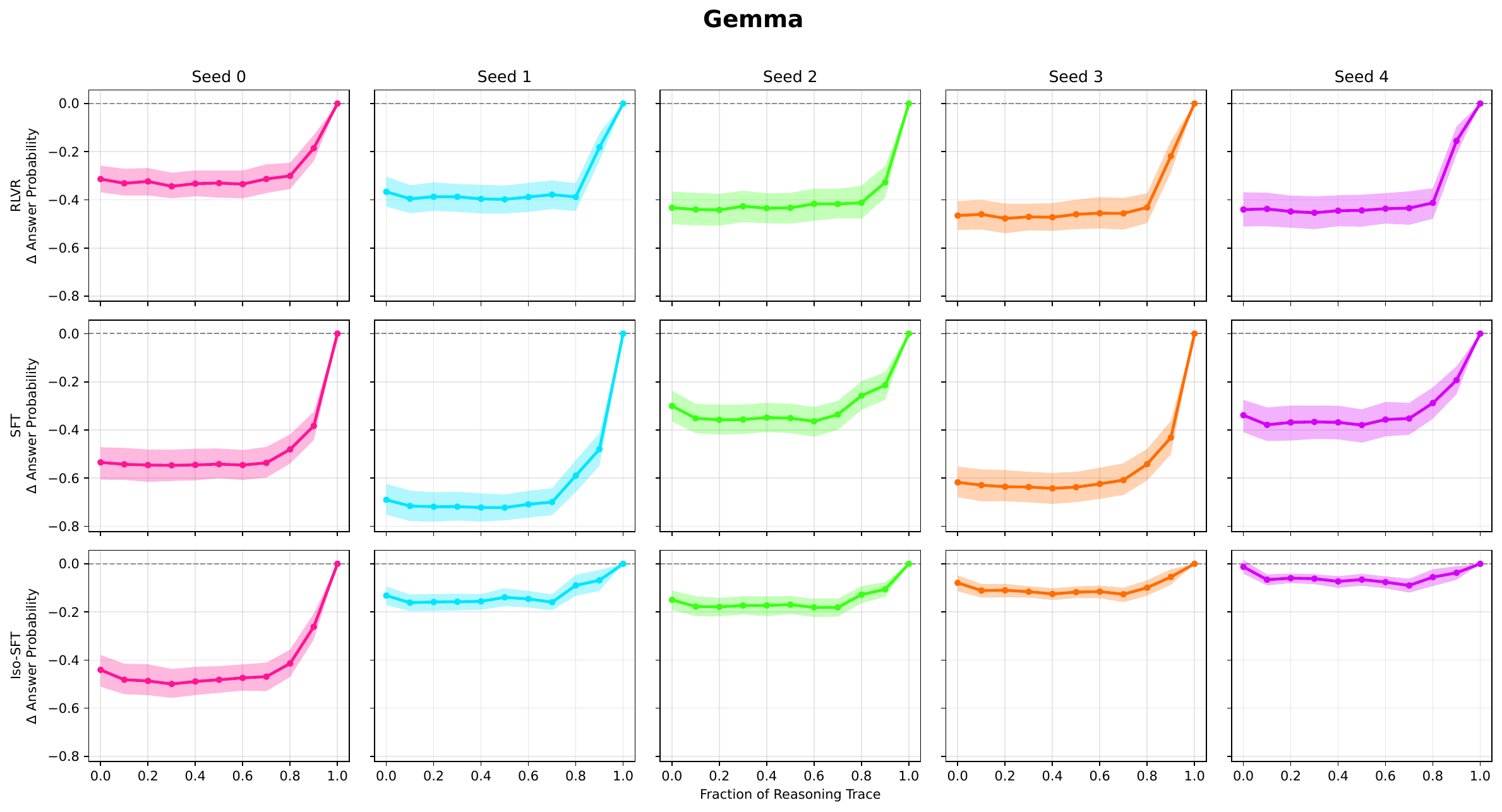}
\includegraphics[width=82mm]{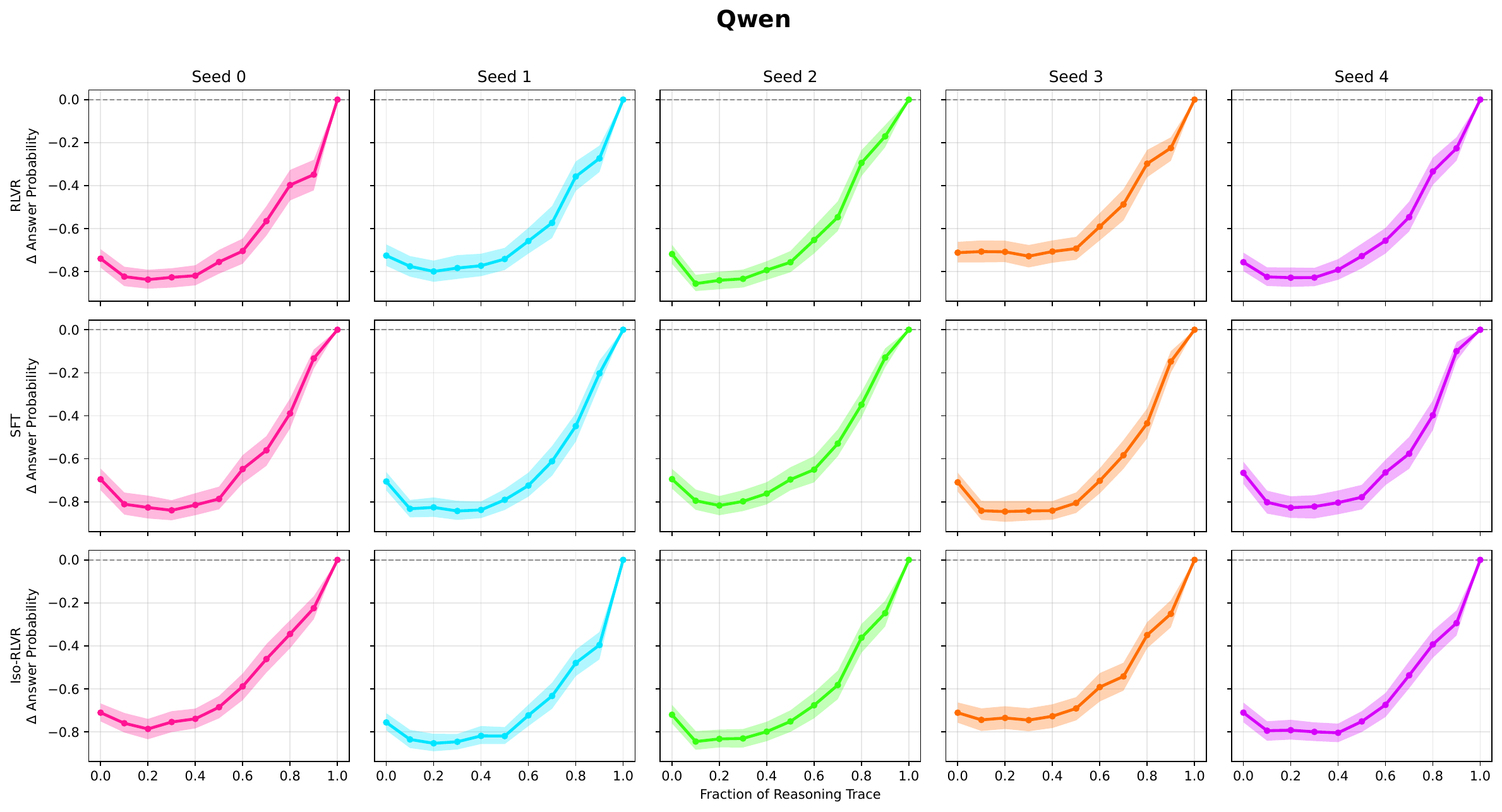}
\end{center}
\caption{Mean difference between final-answer probabilities given partial-trace fractions (for $p\in\{0.0,0.1,\dots,1.0\}$) versus
final-answer probabilities given full traces (95\% confidence intervals). Differences at 1.0 (the full trace) are always 0.0 by design.}
\label{fig_logprob_grid}
\end{figure*}

\begin{table*}[h]
\centering
\scriptsize
\setlength{\tabcolsep}{3pt}
\resizebox{\textwidth}{!}{%
\begin{tabular}{l|l|c|c|c|c|c}
\textbf{Model} & \textbf{Training Type} & \textbf{Seed 0} & \textbf{Seed 1} & \textbf{Seed 2} & \textbf{Seed 3} & \textbf{Seed 4} \\
\hline
\multirow{3}{*}{Llama}
 & RLVR    & $0.6866_{\pm 0.3530}$ & $0.5873_{\pm 0.3509}$ & $0.5205_{\pm 0.3794}$ & $0.6903_{\pm 0.3477}$ & $0.6149_{\pm 0.3642}$ \\
 & SFT     & $0.6887_{\pm 0.3812}$ & $0.7350_{\pm 0.3584}$ & $0.7491_{\pm 0.3627}$ & $0.7762_{\pm 0.3386}$ & $0.7714_{\pm 0.3496}$ \\
 & Iso-SFT & $0.6973_{\pm 0.3979}$ & $0.5807_{\pm 0.3771}$ & $0.2999_{\pm 0.3272}$ & $0.5924_{\pm 0.3751}$ & $0.6650_{\pm 0.3406}$ \\
\hline
\multirow{3}{*}{Gemma}
 & RLVR    & $0.3365_{\pm 0.3317}$ & $0.3965_{\pm 0.3296}$ & $0.4571_{\pm 0.3699}$ & $0.4726_{\pm 0.3438}$ & $0.4549_{\pm 0.3956}$ \\
 & SFT     & $0.5231_{\pm 0.3719}$ & $0.7223_{\pm 0.3226}$ & $0.3363_{\pm 0.3606}$ & $0.6304_{\pm 0.3607}$ & $0.3723_{\pm 0.4095}$ \\
 & Iso-SFT & $0.4721_{\pm 0.3435}$ & $0.1419_{\pm 0.2569}$ & $0.1605_{\pm 0.2534}$ & $0.1185_{\pm 0.1835}$ & $0.0505_{\pm 0.1464}$ \\
\hline
\multirow{3}{*}{Qwen}
 & RLVR     & $0.8970_{\pm 0.2148}$ & $0.8794_{\pm 0.2357}$ & $0.9163_{\pm 0.1606}$ & $0.8390_{\pm 0.2332}$ & $0.8990_{\pm 0.2068}$ \\
 & SFT      & $0.8652_{\pm 0.2434}$ & $0.8844_{\pm 0.2058}$ & $0.8515_{\pm 0.2322}$ & $0.8719_{\pm 0.2271}$ & $0.8458_{\pm 0.2783}$ \\
 & Iso-RLVR & $0.8375_{\pm 0.2351}$ & $0.8754_{\pm 0.2002}$ & $0.8879_{\pm 0.1867}$ & $0.8499_{\pm 0.2560}$ & $0.8601_{\pm 0.2475}$ \\
\end{tabular}%
}
\caption{Mean difference ($\pm$ standard deviation) between final-answer probability given the original CoT and final-answer probability given a swapped CoT. All values in this table are significant at $p<0.001$ under a paired t-test.}
\label{table_cot_swap_grid}
\end{table*}

\subsection{Performative CoT}

We first verified that the models were not generating performative CoTs: i.e.\ determining the answer early on, and simply generating a CoT post-hoc to mimic their training distribution \citep{boppana2026reasoning}. 

Recall that in the trace legibility experiment in Section \ref{sec_experiments_sub_method_sub_evaluation}, we sampled for each query $x$ a trace/answer pair $(z,y)\sim\basemodel\condist[\text{\textendash}]{x}$, then split $z$ into a sequence of partial traces $z_{0.0},\dots,z_{0.9}$. In the present experiment, we measured the probability $\basemodel\condist[y]{x,z_n}$ of the answer $y$ conditioned on each partial trace $z_n$, and compared it to the probability $\basemodel\condist[y]{x,z}$ of $y$ given the full trace $z$: if $\pi$ is generating performative CoTs, we would expect $\basemodel\condist[y]{x,z_n}\approx\basemodel\condist[y]{x,z}$. On the other hand, if $\pi$ is actually using the full reasoning trace, we would expect $\basemodel\condist[y]{x,z_n}\ll\basemodel\condist[y]{x,z}$.

In Figure \ref{fig_logprob_grid}, we plot the mean (over all queries $x$) difference in final-answer probability conditioned on partial traces compared to final-answer probability conditioned on full traces. Across nearly all models and seeds\textemdash with the exception of Gemma Iso-SFT\textemdash we see final-answer probability increasing as the partial trace fraction approaches 1, indicating that the models are in fact using their CoTs.

\subsection{Invisible Reasoning}

Next, we verified that the models were not performing invisible reasoning: simply using the CoT to perform computations in the hidden states to arrive at the final answer, independently of the actual tokens in the CoT \citep{baherwani2026not}.

In this experiment, we collected the 128 trace/answer pairs $(z_i,y_i)\sim\basemodel\condist[\text{\textendash}]{x_i}$, and bucketed the reasoning traces by length. The traces were placed into ten buckets: one for each decile. 

For each query/trace/answer triple $(x_i,z_i,y_i)$, we computed the final-answer probability $\basemodel\condist[y_i]{x_i,z_k}$ conditioned on five $z_k$ from the same length bucket as $z_i$ (ensuring that $k\neq i$), and compared these probabilities to the original $\basemodel\condist[y_i]{x_i,z_i}$. If $\basemodel$ is performing invisible reasoning, we would expect that the actual tokens in the reasoning chain do not impact final-answer probability. Otherwise, we would expect that the probability of $y_i$ conditioned on a reasoning trace $z_k$\textemdash generated for a different query $x_k$\textemdash would be significantly lower than conditioned on the original trace $z_i$: $\basemodel\condist[y_i]{x_i,z_k}\ll\basemodel\condist[y_i]{x_i,z_i}$.

In Table \ref{table_cot_swap_grid}, we report the mean difference between original answer probability and swapped answer probability: this value is positive for all models, indicating that they are not performing invisible reasoning in their CoTs. 

\section{Example Reasoning Traces}
\label{app_examples}

\subsection{Llama}
\label{app_examples_sub_llama}

\begin{figure*}[h!]
\begin{center}
\includegraphics[width=82mm]{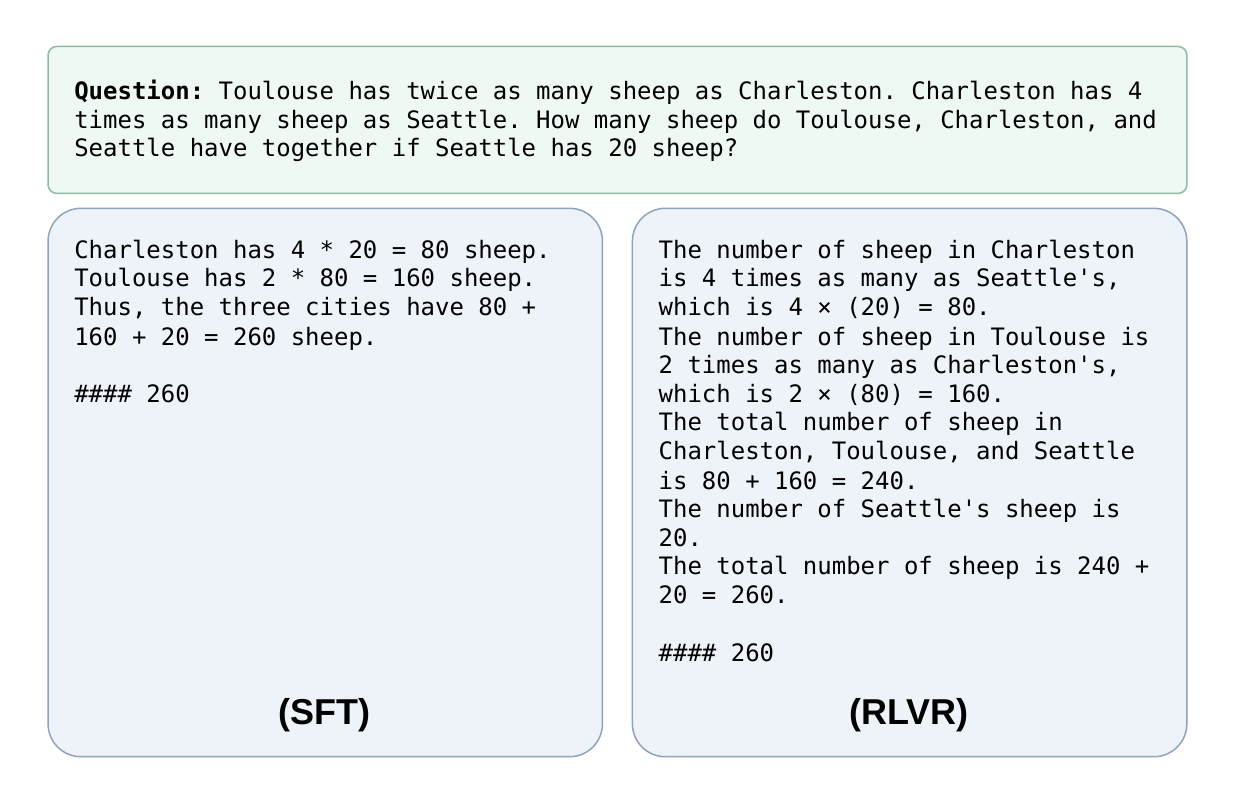}
\includegraphics[width=82mm]{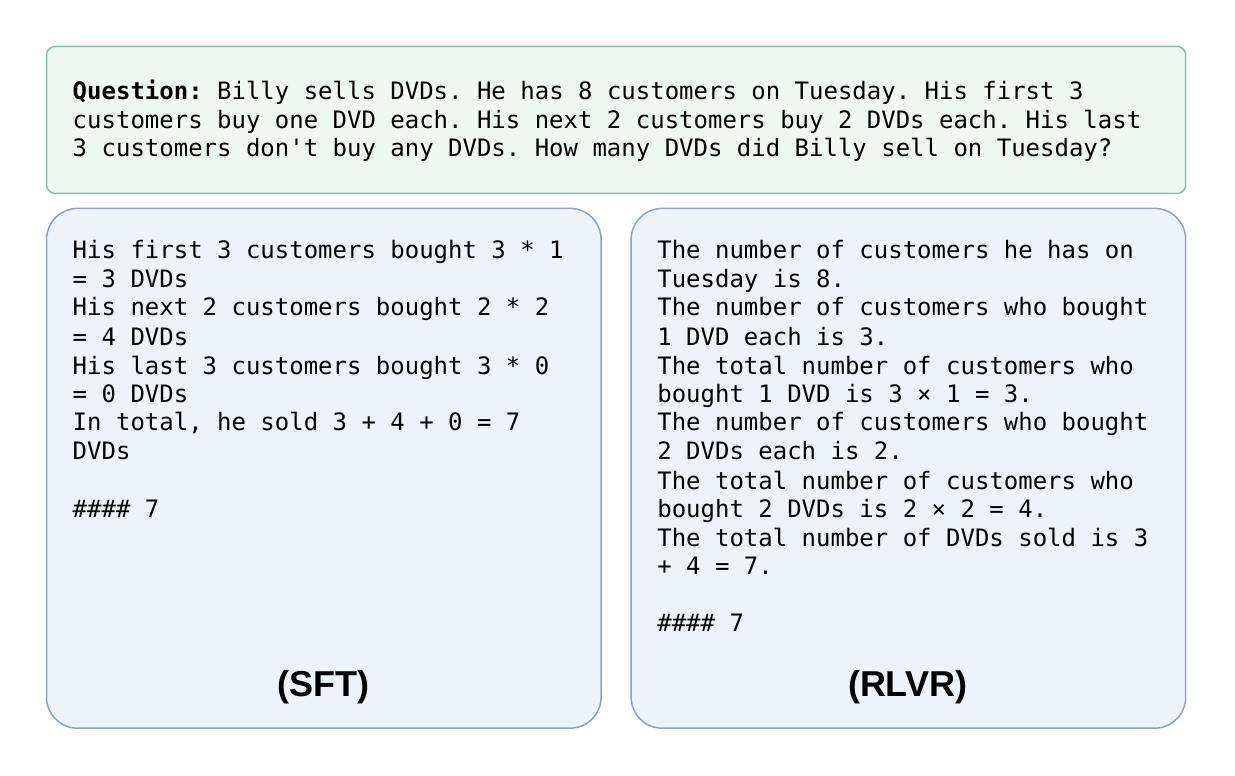}
\end{center}
\end{figure*}

\begin{figure*}[h!]
\begin{center}
\includegraphics[width=82mm]{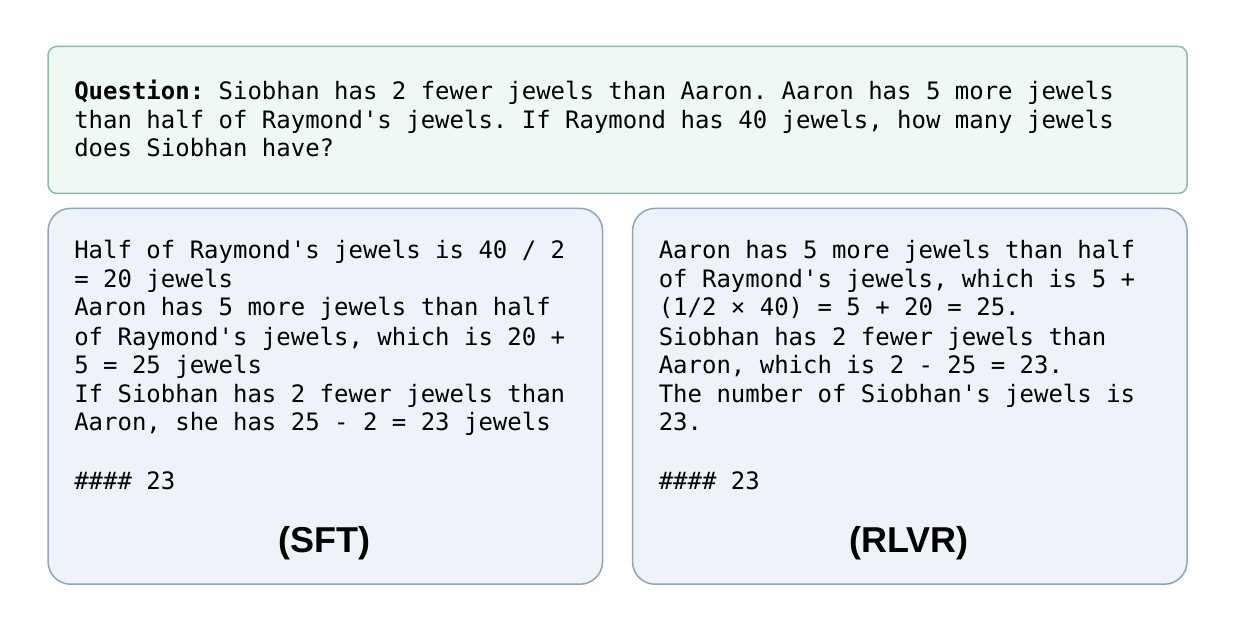}
\includegraphics[width=82mm]{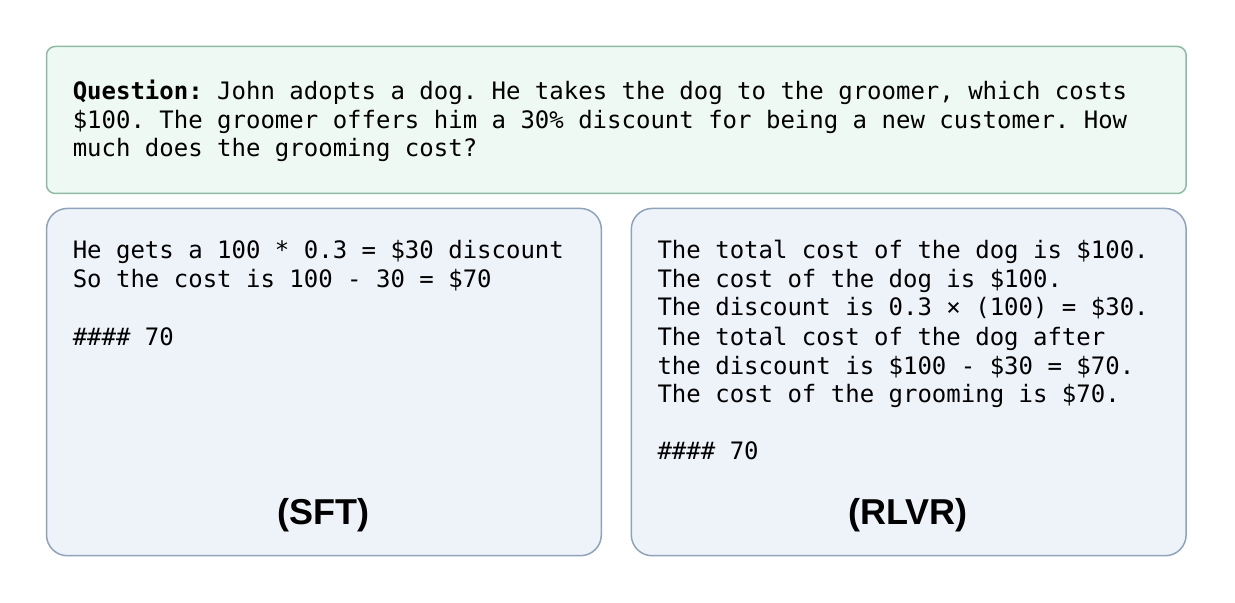}
\end{center}
\end{figure*}

\begin{figure*}[h!]
\begin{center}
\includegraphics[width=82mm]{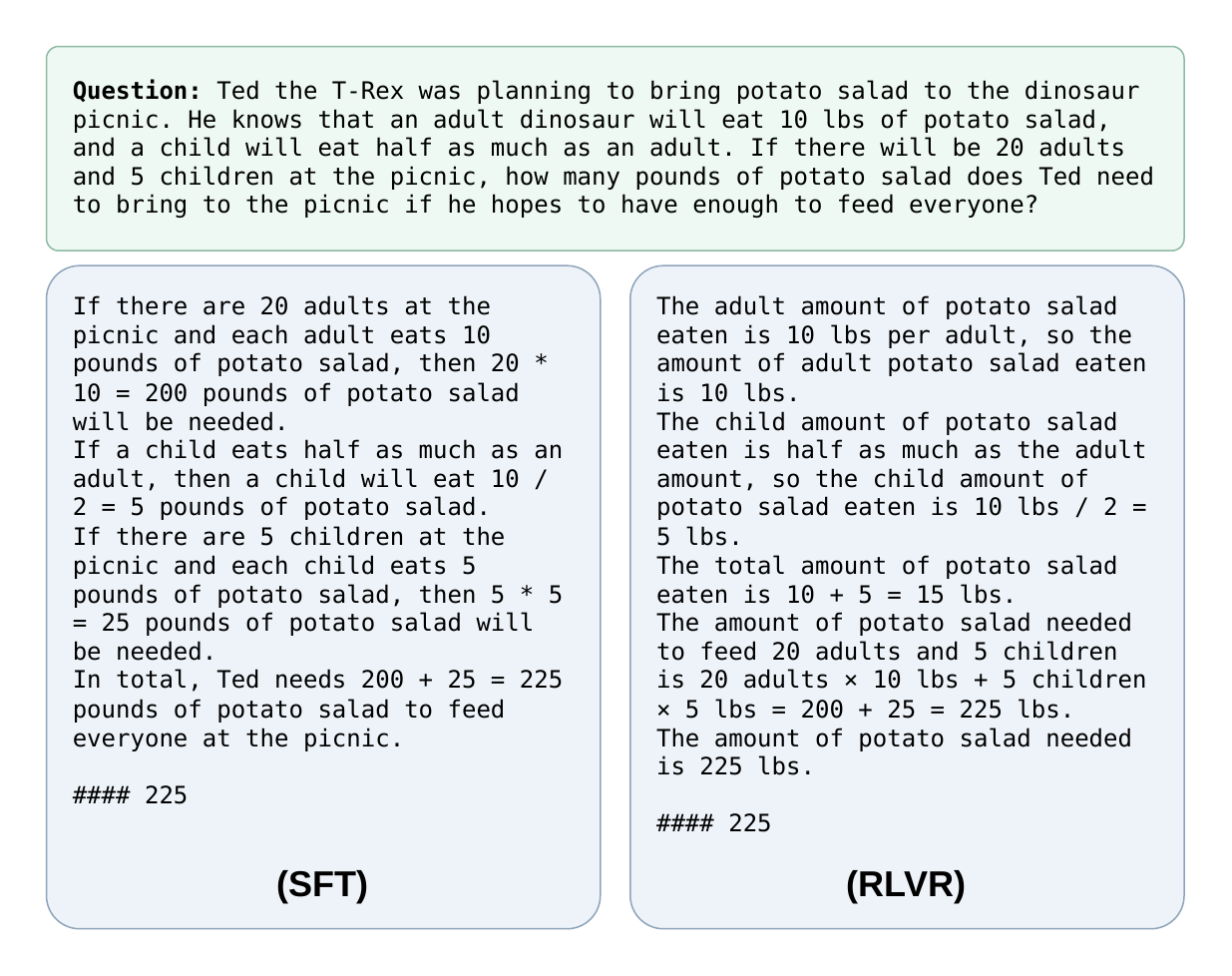}
\includegraphics[width=82mm]{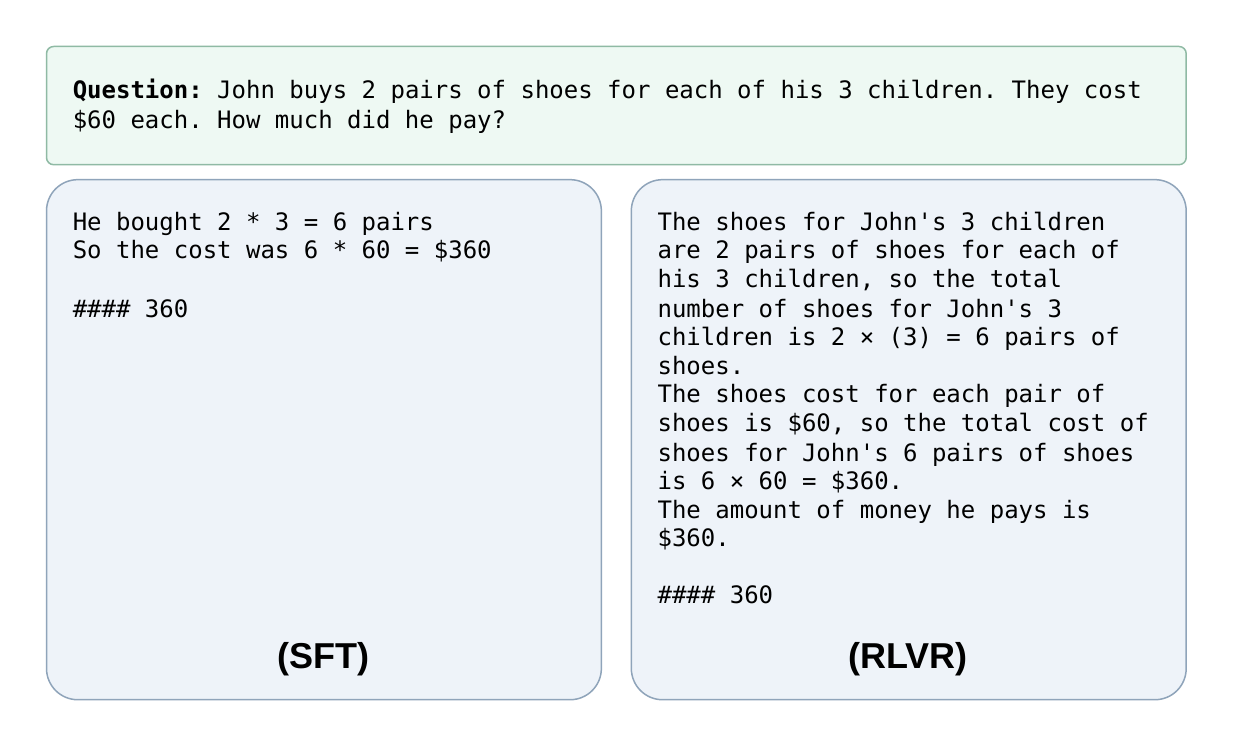}
\end{center}
\end{figure*}

\begin{figure*}[h!]
\begin{center}
\includegraphics[width=82mm]{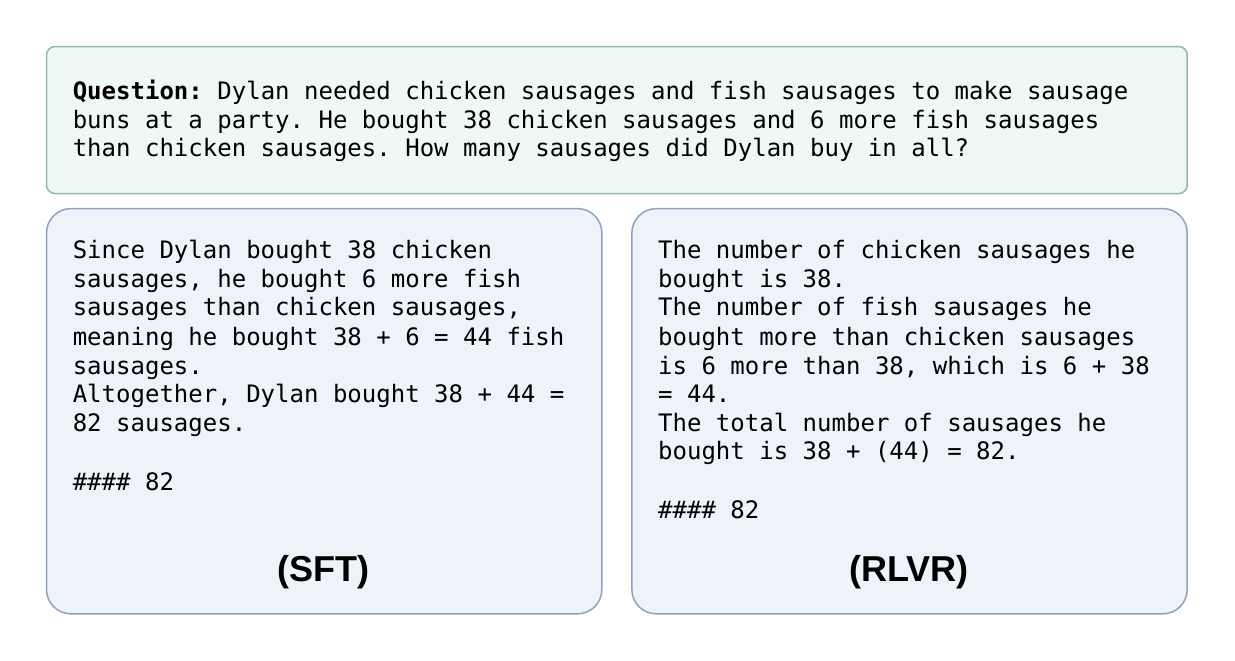}
\end{center}
\end{figure*}

\newpage

\hspace{0mm}

\newpage

\subsection{Gemma}
\label{app_examples_sub_gemma}

\begin{figure*}[h]
\begin{center}
\includegraphics[width=82mm]{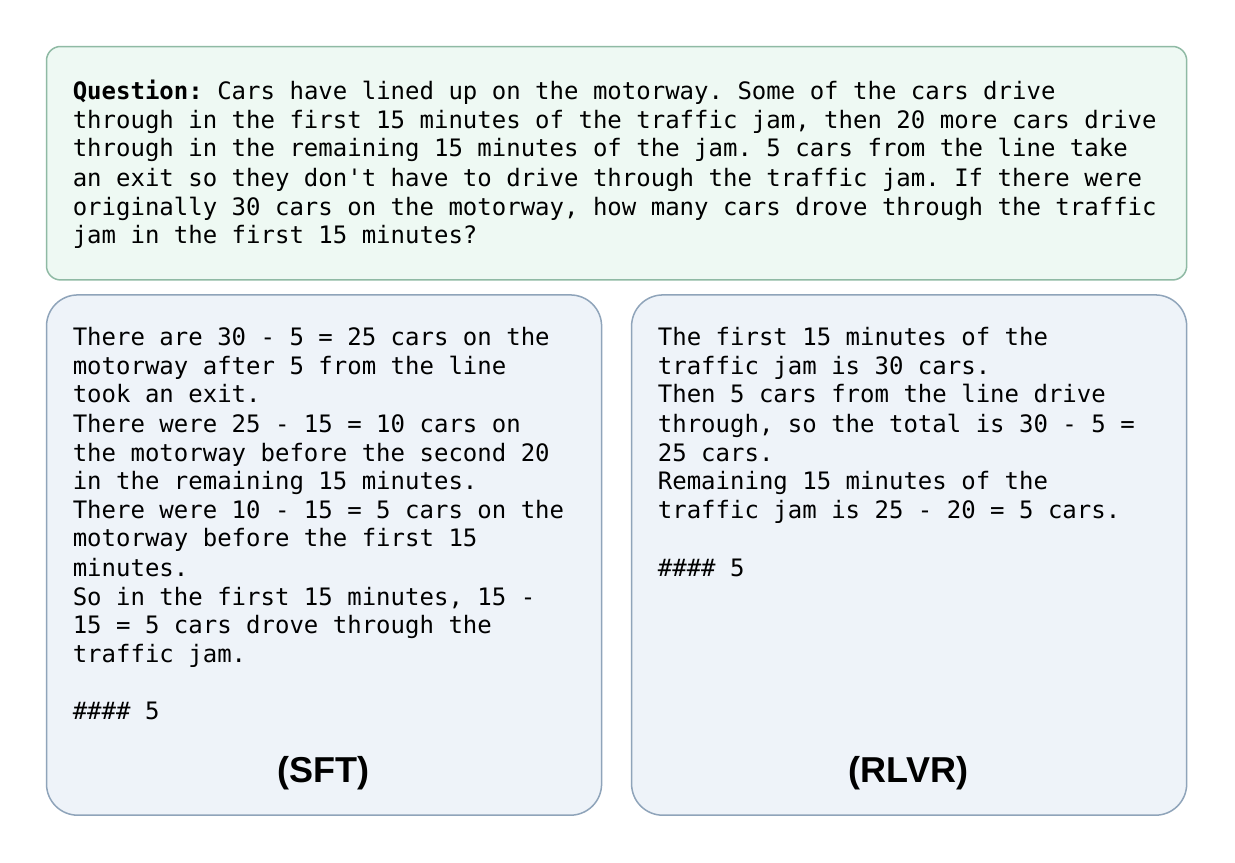}
\includegraphics[width=82mm]{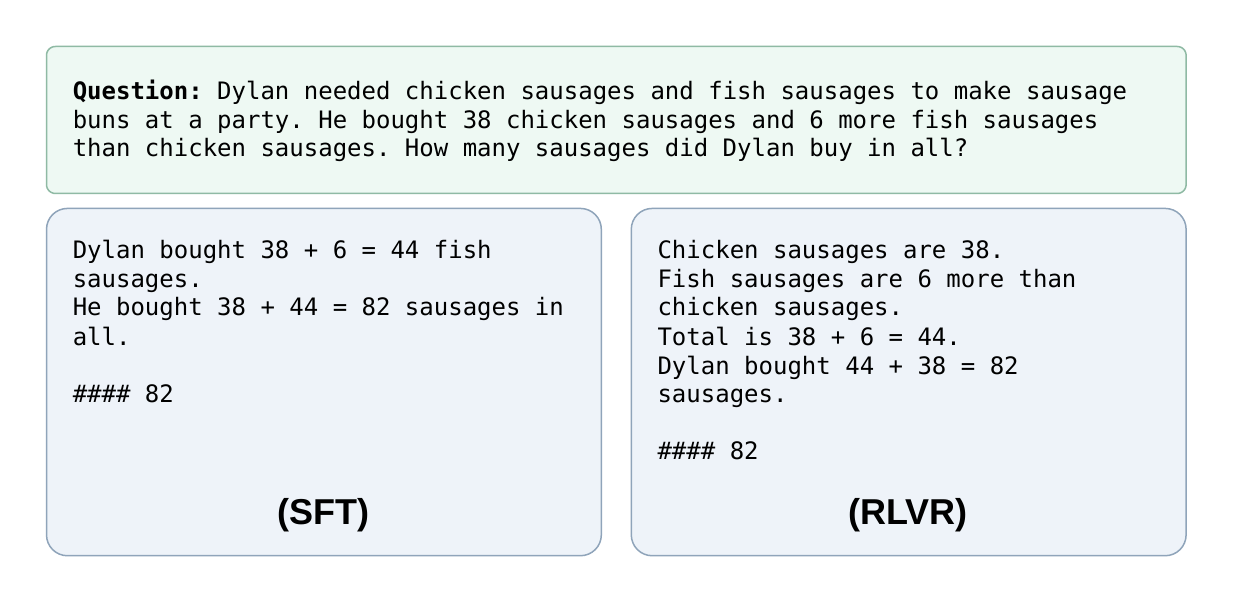}
\end{center}
\end{figure*}

\begin{figure*}[h]
\begin{center}
\includegraphics[width=82mm]{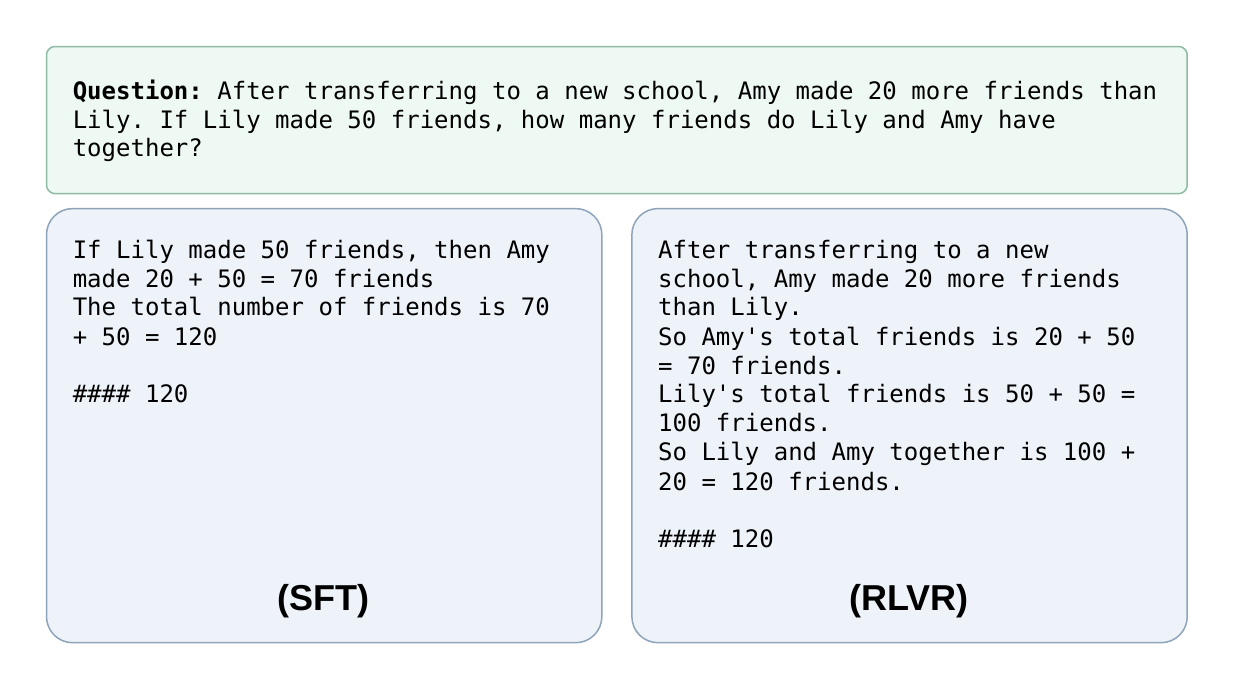}
\end{center}
\end{figure*}

\end{document}